\documentclass[sigconf,nonacm]{acmart}
\setcopyright{none}
\renewcommand\footnotetextcopyrightpermission[1]{}

\usepackage{amsmath}
\usepackage{algorithm}
\usepackage{algorithmic}
\usepackage{xcolor}
\usepackage{color}
\usepackage{fancyvrb}

\DefineVerbatimEnvironment{Highlighting}{Verbatim}{commandchars=\\\{\}}

\usepackage{longtable,booktabs,array}
\usepackage{enumitem}
\usepackage{calc} 
\usepackage{etoolbox}
\makeatletter
\patchcmd\longtable{\par}{\if@noskipsec\mbox{}\fi\par}{}{}
\makeatother
\IfFileExists{footnotehyper.sty}{\usepackage{footnotehyper}}{\usepackage{footnote}}
\makesavenoteenv{longtable}
\usepackage{graphicx}
\makeatletter
\newsavebox\pandoc@box
\newcommand*\pandocbounded[1]{
  \sbox\pandoc@box{#1}%
  \Gscale@div\@tempa{\textheight}{\dimexpr\ht\pandoc@box+\dp\pandoc@box\relax}%
  \Gscale@div\@tempb{\linewidth}{\wd\pandoc@box}%
  \ifdim\@tempb\p@<\@tempa\p@\let\@tempa\@tempb\fi
  \ifdim\@tempa\p@<\p@\scalebox{\@tempa}{\usebox\pandoc@box}%
  \else\usebox{\pandoc@box}%
  \fi%
}
\def\fps@figure{htbp}
\makeatother
\title{Program Distillation: Transferring LLM Capability from System 2 to System 0}
\author{Pujun Zheng}
\affiliation{%
  \institution{School of Economics and Management, East China Normal University}
  \city{Shanghai}
  \country{China}}
\email{pjzheng@stu.ecnu.edu.cn}

\author{Wanying Ren}
\affiliation{%
  \institution{School of Statistics, East China Normal University}
  \city{Shanghai}
  \country{China}}
\email{wyren@stu.ecnu.edu.cn}

\author{Guoxiu He}
\affiliation{%
  \institution{School of Economics and Management, East China Normal University}
  \city{Shanghai}
  \country{China}}
\email{gxhe@fem.ecnu.edu.cn}
\correspondingauthor

\renewcommand{\shortauthors}{Zheng et al.}
\begin{document}

\begin{abstract}
Recent studies leverage large language models (LLMs) to construct executable programs through environment interaction, feedback analysis, and code revision. However, prior work has mainly treated LLMs as optimizers, overlooking their ability to transfer knowledge across system levels. We propose Program Distillation (PD), a framework that formalizes the transfer of abilities from System 2 to System 0. During learning, the LLM provides task knowledge and draft strategies; at deployment, a student program executes the resulting policy. We show that this approach can improve performance while reducing inference time. We introduce PDBench, a benchmark comprising 27 discrete- and continuous-control tasks with fixed initial policies, and PDMetaHarness, a model-harness-independent meta-system for monitoring and guiding the learning process. The strongest model achieves a normalized mean score of 0.728, surpassing PPO (0.656) and Jev-like decision models (0.239), while running exclusively on CPUs and reducing inference costs. Our analysis shows that performance is largely driven by the first breakthrough during learning, and structural modifications have a slightly lower success rate than parameter modifications but typically yield larger gains. We further introduce style distillation, which enables interpretable stylistic control over the learning process. Overall, we present an analytical framework for PD and demonstrate its practical viability.
\end{abstract}


\maketitle
\begingroup
\let\thefootnote\relax
\footnotetext{\textsuperscript{\ding{41}} Corresponding author.}
\endgroup

\section{Introduction}\label{introduction}

\begin{figure}
	\centering
	\pandocbounded{\includegraphics[keepaspectratio]{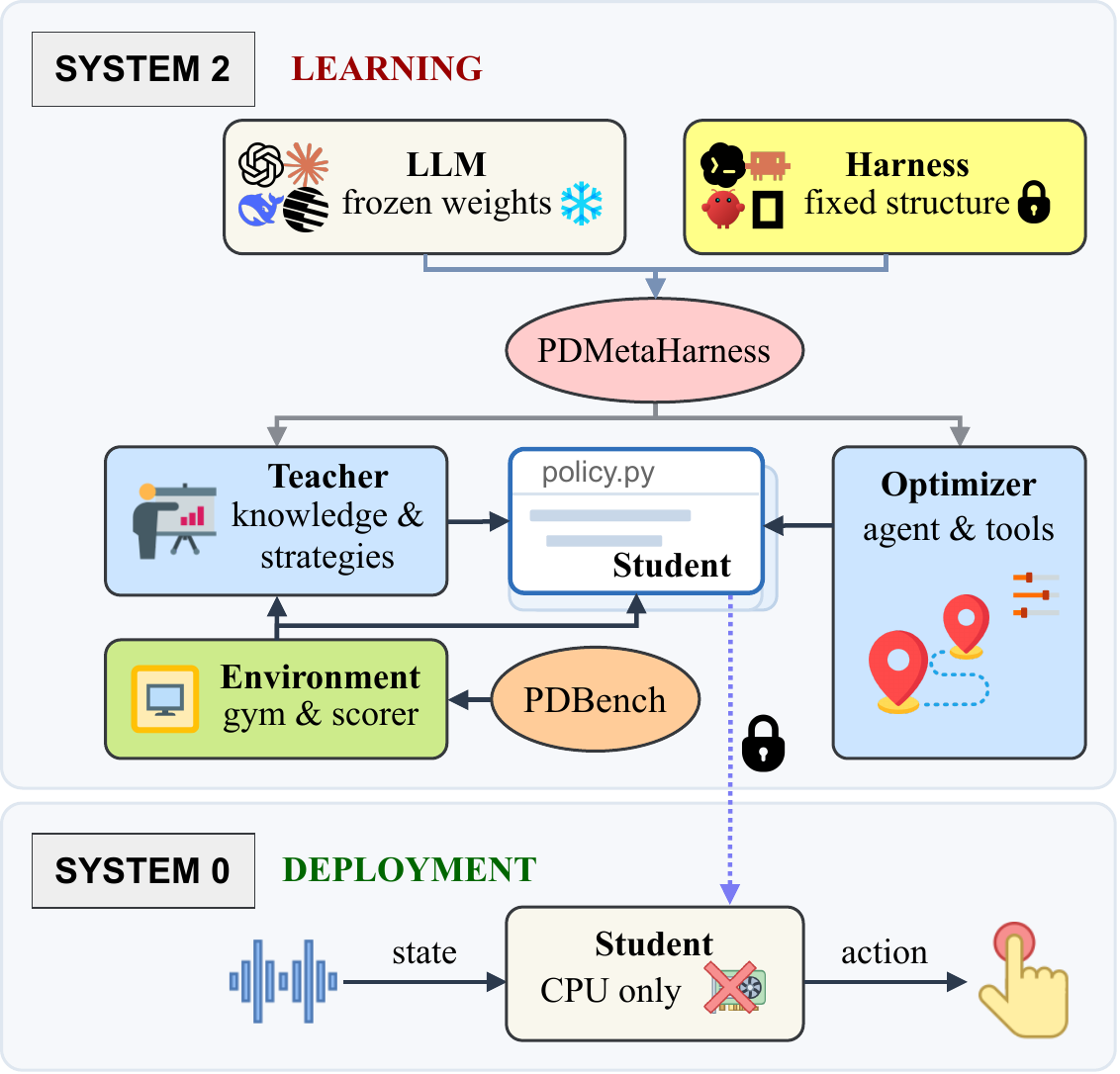}}
	\caption{Program distillation overview. A frozen System 2 LLM serves as both the distillation teacher and the learning optimizer. As a teacher, it supplies task-specific knowledge and uses its parametric knowledge to propose control strategies; as an optimizer, it organizes experiments, invokes tools, and interprets feedback. The environment evaluates candidate programs, and the final program is frozen and deployed as a System 0 policy.}\label{fig:distillation-overview}
\end{figure}

Recent studies have used large language models (LLMs) to construct executable policies \cite{liang2023code}. Given an initial program, an LLM interprets task rules and iteratively develops a policy through environment interaction, feedback analysis, and code revision \cite{liu2024eoh, weng2026learning_beyond_gradients, ishida2024langprop, kuang2025generative, hatgiskessell2026promptpo}. The resulting program encodes task experience in its structure, rules, and parameters, and can be executed in sequential decision-making tasks.

However, prior work has primarily treated LLMs as program optimizers that generate, search for, and improve programs, overlooking their potential to transfer knowledge across system levels. We propose program distillation (PD), which formalizes this process as knowledge distillation from System 2 to System 0. System 2 denotes deliberative, slow thinking, whereas System 0 refers to a pre-cognitive, embodied process that does not require explicit neural processing \cite{taniguchi2025quad}. In conventional knowledge distillation, teacher guidance is used to train a student and improve its capabilities \cite{hinton2015distilling}. Prior work has demonstrated distillation within the same system, including the transfer of decision-making capabilities from System 1 to System 1 \cite{rusu2015policy} and reasoning capabilities from System 2 to System 2 \cite{deepseekai2025r1}. Other approaches transfer capabilities across systems. For example, System 2-to-System 1 distillation trains a model to respond directly based on the outputs of deliberative computation \cite{yu2024system2, chen2026system1arena}. By contrast, PD transfers capabilities from System 2 to System 0. During learning, a System 2 LLM deliberates to develop a task-specific control strategy. At deployment, this strategy is embodied in a task-specific System 0 program, enabling repeated execution with low computational overhead and explicit, interpretable logic. Like knowledge distillation, PD aims to transfer capabilities within or across systems to support efficient deployment and inference on edge devices. More broadly, this perspective aligns with calls to return to System 1 \cite{jev2026systemone}. Both approaches seek high capability, low latency, structured outputs, and reliable execution at low cost.

Figure~\ref{fig:distillation-overview} illustrates our architecture, in which the LLM serves as both teacher and learning optimizer. As a teacher, it contributes task knowledge and draws on its parametric knowledge to propose control strategies. As a learning optimizer, it organizes experiments, invokes tools, and interprets feedback. To support this architecture, we develop two complementary components: PDBench and PDMetaHarness.

PDBench is built on 27 Gym tasks, comprising 11 discrete-action and 16 continuous-action environments. For each task, we specify the available state fields, action bounds, environment configuration, scoring objective, and initial policy program. The model's goal is to improve the initial policy and achieve the highest normalized score. PDMetaHarness is a meta-control system that operates independently of the model and harness. It monitors the learning workspace, automatically steers execution, and provides follow-up guidance. It also enables the LLM to serve as both a distillation teacher and a learning optimizer, thereby ensuring the effective execution of the distillation process. Together, these components allow different models and harnesses to follow a common workflow across environments and optimization objectives, enabling fair comparisons.

We evaluate ten LLMs and five additional harnesses on PDBench across 27 discrete- and continuous-control tasks. The best standard learning configuration achieves a normalized mean score of 0.728, exceeding the score of 0.656 achieved by the PPO baseline using conventional reinforcement learning and the score of 0.239 achieved by Jev-like decision models. It also achieves this performance in less time than all non-PD baselines when inference is performed exclusively on a CPU, demonstrating that PD is a more effective and cost-efficient alternative. 

Trajectory analysis shows that performance is largely driven by the first breakthrough during learning. Structural modifications have a slightly lower success rate than parameter modifications, but yield greater gains. We also introduce a distillation strategy called style distillation, which injects interpretable, stylized guidance during distillation. Our results further suggest that some tasks may have a unique optimal policy. Overall, these results demonstrate the practical viability of PD.

Our contributions are as follows:
\begin{enumerate}[leftmargin=*]
\item \textbf{Knowledge distillation interpretation.} We formulate PD as knowledge distillation, with a System 2 LLM as the teacher and a System 0 program as the student.
\item \textbf{Infrastructure for PD training and evaluation.} PDBench enables fair comparisons of models, base harnesses, and learning-control policies. PDMetaHarness coordinates the learning process and maintains effective system execution through adaptive guidance.
\item \textbf{Cross-system leading performance.} We show that PD achieves the highest performance while requiring the shortest inference time compared with conventional RL and Jev-like decision models.
\item \textbf{PD optimization strategies and style distillation.} We introduce optimization strategies that further advance the performance frontier, as well as a style-distillation approach for interpretable control of model behavior.
\item \textbf{Trajectory insights from large-scale learning.} We show how performance gains in PD relate to the first breakthrough and structural modifications during learning.
\end{enumerate}

\section{Methodology}\label{methodology}

\subsection{Program Distillation}\label{program-distillation}

We formulate PD as a form of capability distillation from System 2 to System 0 \cite{taniguchi2025quad}. An LLM with fixed parameters serves as the teacher, and a program serves as the student. Through task-specific reasoning and experimentation in the environment, it transfers task-relevant execution capabilities into standalone decision rules that can be deployed on edge devices. Complex reasoning is concentrated in the learning phase, while the student acts continuously at deployment time with low online computational cost. PD thus involves two related transformations. First, it changes the representation of capability: knowledge in an LLM is converted into the structure and parameters of a specialized program. Second, it changes the mode of execution: decisions that require online model inference are replaced by direct execution of a local program. 

\paragraph{Correspondence to conventional distillation.} Given a teacher \(T_\theta\), a student \(S_\phi\), and transfer data \(\mathcal D\), a common distillation objective accounts for both task supervision and teacher guidance \cite{hinton2015distilling}:

\begin{equation}
\phi^\star\in\arg\min_\phi
\left[
\mathcal L_{\mathrm{task}}(S_\phi;\mathcal D)
+\lambda\mathcal L_{\mathrm{dist}}(S_\phi,T_\theta;\mathcal D)
\right].
\end{equation}

Here, \(\mathcal L_{\mathrm{task}}\) denotes task supervision, \(\mathcal L_{\mathrm{dist}}\) denotes teacher guidance, and \(\lambda\geq 0\) controls the trade-off between them. The teacher parameters \(\theta\) are fixed, whereas the student parameters \(\phi\) are learnable. In PD, the student is a fixed policy program \({S_\phi}\). Outcomes from the environment provide task supervision, while the teacher draws on its existing knowledge to reason about effective rules and code, then revises the student's policy program. The overall principle is similar to that of conventional distillation.

\paragraph{System capability and speed.} We distinguish two dimensions in our objective: decision capability, which describes how well a system performs a task, and execution speed, which measures the time from receiving an observation to producing a usable action.

Let the task define a metric for which higher values are better, and let \(f\) denote either the teacher \(T\) or the student \(S\). We define

\begin{equation}
C(f)=\mathbb E_{\xi}
\left[\mathsf{Score}_{\mathcal T}(\tau^{(0)}_{f,\xi})\right].
\end{equation}

Here, \(\tau^{(0)}_{f,\xi}\) is an idealized model outcome under seed \(\xi\), assuming the task is not time-sensitive. Thus, \(C(f)\) represents the system's decision capability under this idealized setting.

In practical PD applications, however, many systems are time-sensitive, and the inference latency of most deployed models is constrained. We therefore account for a factor \(\Delta\) that affects system performance. Actual execution performance is determined by rollout on the real system:

\begin{equation}
\begin{aligned}
\tau^{(\Delta)}_{f,\xi}
&=\mathsf{Rollout}(E,f;\xi),\\
J_\Delta(f)&=\mathbb E_{\xi}
\left[\mathsf{Score}_{\mathcal T}(\tau^{(\Delta)}_{f,\xi})\right]=C(f)-\Lambda_\Delta(f).
\end{aligned}
\end{equation}

Here, \(\Lambda_\Delta(f)\) is a time-related penalty that represents the gap between actual and ideal performance. Accordingly, \(J_\Delta(f)\) measures the model's performance in the real environment.

Finally, we define the following metric for execution speed:

\begin{equation}
v_\Delta(f)=\frac{1}{\ell^{(\Delta)}_{f,\xi}}.
\end{equation}

Here, \(\ell^{(\Delta)}_{f,\xi}\) is the end-to-end inference time for rollout on the real system, and \(v_\Delta(f)\) is its reciprocal.

Notably, in time-sensitive decision problems, a student can trade some decision capability for lower response latency. When the benefit of faster responses outweighs the capability gap, deployment performance can exceed that of the online teacher. For example, the difference in actual performance between the teacher and the student can be written as

\begin{equation}
J_\Delta(T)-J_\Delta(S)
=\left[C(T)-C(S)\right]
-\left[\Lambda_\Delta(T)-\Lambda_\Delta(S)\right].
\end{equation}

The left-hand side represents the teacher's deployment advantage over the student. On the right-hand side, the first term represents the capability gap, and the second represents the temporal gap. When the impact of timing exceeds the capability gap, this expression is negative, yielding \(J_\Delta(S)>J_\Delta(T)\). In this case, the student outperforms the teacher. If we also observe \(v_\Delta(S)>v_\Delta(T)\), the program student achieves better deployment performance through faster responses.

\subsection{Learning Process}\label{feedback-driven-learning-process}

To perform program distillation in a general manner that is independent of the model and harness framework, we design PDMetaHarness, a meta-control system defined as \(\mathcal K\).

\paragraph{Design of PDMetaHarness.} PDMetaHarness is defined over a model \(M_\theta\) and a harness \(H\). It guides the model \(M_\theta\) to act as both the teacher \(T_\theta\) and the optimizer \(\mathcal O_\theta\).

\begin{equation}
(T_\theta,\mathcal O_\theta)
=\mathcal K(M_\theta,H).
\end{equation}

\(M_\theta\) is an LLM with fixed parameters. The base harness \(H\) handles model calls, tool interactions, and session execution. The meta-control rules \(\mathcal K\) comprise common user control instructions, Steering and Followup. In our setting, PDMetaHarness triggers these automatically based on feedback from the learning workspace context \(C\). In practical applications, it can handle issues such as stalled progress and premature termination without user involvement. Concretely, \(\mathsf{Steer}(C)\) monitors the workspace during each distillation step and can inject contextual prompts when learning stalls, whereas \(\mathsf{Follow}(C)\) reopens the run with the existing context after the model stops automatically.

\paragraph{Program distillation objective.} Let \(M_\theta\) denote the frozen LLM, \(t\in\mathcal T\) a task, and \(E\) the environment. Starting from \({S_\phi}_0\), the teacher proposes modifications, and the optimizer uses development feedback to decide which modifications to integrate and how to refine them. The teacher's weights remain fixed, while the program and workspace context are updated continuously. The meta-harness orchestrates the learning loop: the teacher generates candidate programs, and the benchmark evaluates their task performance.

During distillation, the program first generates execution trajectories, and the evaluator then computes the mean task score over a fixed set of seeds:

\begin{equation}
\begin{aligned}
\tau_{{S_\phi},\xi}
&=\mathsf{Rollout}(E,{S_\phi};\xi),\\
\widehat J_q({S_\phi})
&=\frac{1}{|D_q|}\sum_{\xi\in D_q}
\mathsf{Score}_{\mathcal T}(\tau_{{S_\phi},\xi}).
\end{aligned}
\label{eq:def-j}
\end{equation}

Here, \(\tau_{{S_\phi},\xi}\) is the complete trajectory under seed \(\xi\), and \(q\in\{\mathrm{dev},\mathrm{test}\}\) denotes development or test evaluation. Development and test evaluation use independent seed sets \(D_{\mathrm{dev}}\) and \(D_{\mathrm{test}}\).

For the maximization tasks considered in this paper, the learning objective is

\begin{equation}
{S_\phi}^*\in\arg\max_{{S_\phi}}
\widehat J_{\mathrm{dev}}({S_\phi}),
\end{equation}

where \({S_\phi}^*\) is the optimal control policy.
Algorithm~\ref{alg:pdmetaharness} details the complete procedure for executing this learning strategy.

\begin{algorithm}[t]
	\caption{Feedback-Driven Learning Process}
	\label{alg:pdmetaharness}
	\begin{algorithmic}[1]
		\small
		\REQUIRE Initial policy ${S_\phi}_0$,
		candidate budget $B$, environment $E$, initial context $C_0$, steering rule $\mathsf{Steer}$,
		follow-up rule $\mathsf{Follow}$.
		\ENSURE Final policy ${S_\phi}^*$, test score $J^*$
		\STATE $C\gets C_0$, ${S_\phi}\gets{S_\phi}_0$, $J\gets\widehat J_{\mathrm{dev}}({S_\phi})$, $k\gets0$
		\WHILE{$k<B$}
		\STATE $S_\phi^{'} = \mathsf{Distill}\left(S_\phi, T_\theta,E,C; \mathcal O_\theta\right)$, with $\mathsf{Steer}(C)$ during execution
		\STATE $J'\gets\widehat J_{\mathrm{dev}}({S_\phi}')$
		\IF{${S_\phi}'$ is valid \AND $J'>J$}
		\STATE Freeze ${S_\phi}'$ as the current best:
		${S_\phi}\gets{S_\phi}'$, $J\gets J'$
		\ENDIF
		\STATE $k\gets k+1$
		\STATE $C\gets\mathsf{Follow}(C)$
		\ENDWHILE
		\STATE ${S_\phi}^*\gets{S_\phi}$
		\STATE $J^*\gets\widehat J_{\mathrm{test}}({S_\phi}^*)$
		\RETURN ${S_\phi}^*$, $J^*$
	\end{algorithmic}
\end{algorithm}

\subsection{Benchmark Design}\label{benchmark-design}

To ensure that model optimization and evaluation are conducted under fair conditions, we design PDBench, a standalone benchmark tailored to program distillation.

\paragraph{Design of PDBench.} PDBench is built on 27 publicly available classic Gym tasks \cite{towers2024gymnasium}. The tasks cover rule selection, path planning, feedback control of posture, target control, and multi-joint coordination. They include 11 discrete-action tasks and 16 continuous-action tasks. Each task is initialized with a valid, unoptimized initial policy. The model can access the specified state fields, action ranges, environment configuration, time limits, and scoring objectives, and iteratively optimize the policy based on this information.

\begin{table}[t]
\centering
{\footnotesize
\caption{Task coverage of PDBench. Each of the 27 tasks is assigned to exactly one primary semantic category; counts and representative tasks summarize the five groups.}\label{tab:task-coverage}
\par\smallskip
\setlength{\tabcolsep}{4pt}
\begin{tabular}{@{}
  >{\raggedright\arraybackslash}p{(\linewidth - 6\tabcolsep) * \real{0.22}}
  >{\raggedleft\arraybackslash}p{(\linewidth - 6\tabcolsep) * \real{0.10}}
  >{\raggedright\arraybackslash}p{(\linewidth - 6\tabcolsep) * \real{0.30}}
  >{\raggedright\arraybackslash}p{(\linewidth - 6\tabcolsep) * \real{0.38}}@{}}
\toprule
\begin{minipage}[b]{\linewidth}\raggedright
Category
\end{minipage} & \begin{minipage}[b]{\linewidth}\raggedleft
Count
\end{minipage} & \begin{minipage}[b]{\linewidth}\raggedright
Representative tasks
\end{minipage} & \begin{minipage}[b]{\linewidth}\raggedright
Primary decision requirement
\end{minipage} \\
\midrule
Game scoring and rule-based decisions & 2 & Pong, Blackjack & Adversarial control and rule selection under hidden information \\
Map navigation and safe paths & 5 & Taxi, FrozenLake, CliffWalking & Path planning and risk management on known maps \\
Postural stability and swinging & 6 & CartPole, Acrobot, HumanoidStandup & Balancing, swinging up, and postural recovery \\
Goal reaching and manipulation & 6 & MountainCar, LunarLander, Pusher & Reaching goals, landing, and object manipulation \\
Sustained locomotion and propulsion & 8 & BipedalWalker, Hopper, Ant, Humanoid & Joint coordination and sustained locomotion \\
Total & 27 & --- & Discrete and continuous control \\
\bottomrule
\end{tabular}}
\end{table}

\paragraph{Score normalization.} Raw returns differ in scale across tasks. To enable meaningful comparisons, we use normalized scores to measure model performance:

\begin{equation}
\mathrm{Score}=\frac{1}{N}\sum_{t=1}^{N}\widetilde J_t,
\qquad
\widetilde J_t=\operatorname{clip}\!\left(
\frac{J_t-a_t}{b_t-a_t},0,1\right).
\end{equation}

Here, \([a_t,b_t]\) is the valid score range, and \(J_t\) is the raw test score of the frozen policy.

\section{Experiment}\label{performance-results-and-learning-control-comparisons}

\subsection{Experimental Setup}\label{experimental-setup}

\paragraph{Baseline Methods}
We evaluate three categories of methods under the same task contract and normalized scoring. (1) \emph{Program distillation (PD)}: we run ten LLM configurations with MiniSWEAgent as the default harness, and in a harness-replacement study we hold GLM-5.3 fixed while varying the harness across OpenClaw, OpenCode, ClaudeCode, OpenHands, and Codex. This category covers standard PD, teacher replacement, and harness replacement. (2) \emph{Jev-like decision models}: SemIf-4B \cite{semif_openjev} and Laya \cite{laya} produce an action at every environment step without within-task policy learning. (3) \emph{Reinforcement learning}: PPO \cite{schulman2017ppo}, SPO \cite{xie2025spo}, RPO \cite{gan2024rpo}, and A3C \cite{mnih2016a3c} train neural policies through environment interaction and parameter updates. The same-model PD category spans ten configurations from multiple providers to reduce dependence on any single provider. Table~\ref{tab:main-results} consolidates the suite-level results for all three categories and reports the timing details.

\paragraph{Basic Configurations}
PD executes at most 30 candidates per iteration and at most 5 standard follow-ups, retaining the best-performing strategy according to development-set performance. The Classic RL baseline uses an MLP, with 2 hidden layers of 64 tanh units in each network. All reinforcement learning experiments use training seed 42 and run for at most 1,000,000 steps per task. The checkpoint with the best development-set performance is retained as the model. The Jev-like model is not trained and is run using the officially released weights. Training of the PD model calls the LLM API, and testing is conducted on a single Intel Core i7-14700KF CPU. Training and inference for Classic RL, as well as inference for Jev-like, are conducted on a single RTX 4070 Ti Super GPU.

\subsection{\emph{RQ1}: How Does PD Compare with Other Learning Paradigms?}\label{main-results}

\begin{figure}
	\centering
	\pandocbounded{\includegraphics[keepaspectratio]{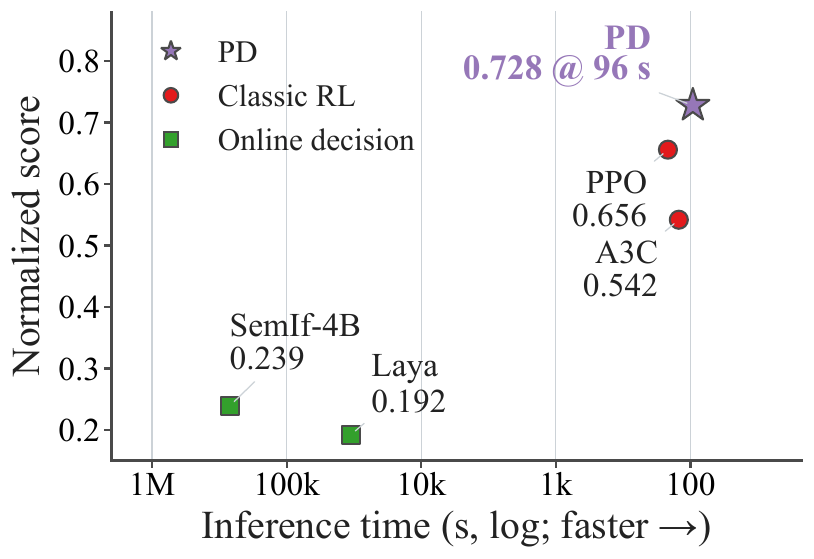}}
	\caption{Normalized score versus inference time for the program distillation, reinforcement-learning, and online-decision routes. The best PD configuration attains a normalized score of 0.728 with shorter inference time than PPO, A3C, SemIf-4B, and Laya. The horizontal axis is logarithmic and runs from slower to faster.}\label{fig:score-inference-time}
\end{figure}

The main results are shown in Table~\ref{tab:main-results} and the Pareto frontier across paradigms is summarized in Figure~\ref{fig:score-inference-time}. Jev-like methods perform poorly: SemIf-4B and Laya obtain overall scores of only 0.239 and 0.192, respectively. Classic reinforcement learning is substantially stronger, with PPO achieving the best overall score among the conventional RL baselines at 0.656. In contrast, the best program distillation (PD) configuration reaches 0.728, exceeding PPO by 0.072. This gap indicates that the main advantage of PD comes from more effective interaction and program-construction strategies rather than from policy optimization alone.

\begin{table}[t]
\centering
{\small
\caption{Performance comparison across the three baseline categories. The best and second-best values across the entire table are shown in bold and underlined, respectively. Build time is cumulative learning-session time. Inference time is the total testing duration for the entire batch; -- denotes not applicable.}\label{tab:main-results}
\par\smallskip
\setlength{\tabcolsep}{2pt}
\begin{tabular}{@{} 
  >{\raggedright\arraybackslash}p{(\linewidth - 10\tabcolsep) * \real{0.32}}
  >{\raggedleft\arraybackslash}p{(\linewidth - 10\tabcolsep) * \real{0.135}}
  >{\raggedleft\arraybackslash}p{(\linewidth - 10\tabcolsep) * \real{0.135}}
  >{\raggedleft\arraybackslash}p{(\linewidth - 10\tabcolsep) * \real{0.135}}
  >{\raggedleft\arraybackslash}p{(\linewidth - 10\tabcolsep) * \real{0.135}}
  >{\raggedleft\arraybackslash}p{(\linewidth - 10\tabcolsep) * \real{0.14}}@{}}
\toprule
\begin{minipage}[b]{\linewidth}\raggedright
Method
\end{minipage} & \begin{minipage}[b]{\linewidth}\raggedleft
Cont. control (16)
\end{minipage} & \begin{minipage}[b]{\linewidth}\raggedleft
Discrete control (11)
\end{minipage} & \begin{minipage}[b]{\linewidth}\raggedleft
Overall (27)
\end{minipage} & \begin{minipage}[b]{\linewidth}\raggedleft
Build time (h)
\end{minipage} & \begin{minipage}[b]{\linewidth}\raggedleft
Infer\-ence time (s)
\end{minipage} \\
\midrule
\multicolumn{6}{@{}l}{\textbf{Jev-like}} \\
SemIf-4B & 0.240 & 0.237 & 0.239 & -- & 264960 \\
Laya & 0.175 & 0.217 & 0.192 & -- & 33480 \\
\addlinespace
\multicolumn{6}{@{}l}{\textbf{Classic RL}} \\
PPO & 0.603 & 0.733 & 0.656 & 5.1 & 147 \\
SPO & 0.558 & 0.643 & 0.593 & 5.1 & 146 \\
RPO & 0.608 & 0.529 & 0.576 & 5.6 & 164 \\
A3C & 0.490 & 0.616 & 0.542 & 9.9 & 122 \\
\addlinespace
\multicolumn{6}{@{}l}{\textbf{PD} \textit{(MiniSWEAgent)}} \\
Claude-Opus-5 & \textbf{0.633} & 0.864 & \textbf{0.728} & 23.0 & 96 \\
MiMo-V2.6-Pro & 0.608 & \textbf{0.884} & 0.720 & 67.3 & 97 \\
GLM-5.3 & \underline{0.618} & 0.864 & 0.718 & 38.0 & 110 \\
DeepSeek-V4.1-Flash & 0.597 & 0.814 & 0.686 & 13.4 & 88 \\
GLM-5.3-Flash & 0.563 & 0.807 & 0.662 & 31.6 & 59 \\
GPT-6-Sol & 0.524 & 0.846 & 0.655 & 16.6 & 74 \\
Claude-Sonnet-5 & 0.541 & 0.780 & 0.638 & 17.1 & 65 \\
Kimi-K2.6 & 0.448 & 0.778 & 0.582 & 34.2 & 73 \\
Qwen3.8-27B & 0.388 & 0.696 & 0.513 & 9.6 & 55 \\
Qwen3.6-35B-A3B & 0.343 & 0.588 & 0.443 & 11.8 & 62 \\
\addlinespace
\multicolumn{6}{@{}l}{\textbf{PD} \textit{(GLM-5.3)}} \\
OpenClaw & 0.613 & \underline{0.881} & \underline{0.722} & 13.4 & 560 \\
OpenCode & 0.611 & 0.843 & 0.706 & 13.4 & 127 \\
ClaudeCode & 0.580 & 0.821 & 0.678 & 11.7 & 76 \\
OpenHands & 0.560 & 0.845 & 0.676 & 9.3 & 105 \\
Codex & 0.374 & 0.600 & 0.466 & 1.8 & 78 \\
\bottomrule
\end{tabular}}
\end{table}

\subsection{\emph{RQ2}: How Do LLM \(M_\theta\) and Harness \(H\) Affect Student \(S_\phi\) Performance?}\label{model-harness-effects}

Across the ten model configurations in Table~\ref{tab:main-results}, Claude-Opus-5 achieves the highest overall score of 0.728, followed by MiMo-V2.6-Pro and GLM-5.3 with scores of 0.720 and 0.718, respectively. All three substantially outperform the unoptimized initial policy, whose average score is only 0.204. The gains are also differentiated by action type: Claude-Opus-5 performs best on continuous control with a score of 0.633, whereas MiMo-V2.6-Pro leads on discrete control with 0.884. This specialization suggests that model capability is not uniformly transferred across action spaces, and that the strongest overall configuration is determined by the balance between continuous execution and discrete decision-making.

The harness comparison further shows that the execution substrate is a major determinant of performance. With GLM fixed, changing the base harness produces overall scores ranging from 0.466 to 0.722. OpenClaw achieves the best overall score of 0.722 and the strongest discrete-control score of 0.881, while MiniSWEAgent obtains the best continuous-control score of 0.618 and a comparable overall score of 0.718. The large spread across harnesses indicates that tool availability and session execution procedures can be as important as the underlying model when constructing executable programs.

\subsection{\emph{RQ3}: How Do Meta-Control Methods \(\mathcal K\) Affect Student \(S_\phi\) Performance?}\label{ablation-study}

\paragraph{Other Control Methods}

We fix the model to GLM-5.3 in the ablation study and use MiniSWEAgent to compare four alternative meta-harness designs.

\textbf{Meta: Cross-task experience transfer.} Reusable experience is extracted from the development checkpoints of completed tasks and injected as a skill into the context of the target task.

\textbf{History: Historical policy comparison.} Each candidate is paired in turn with all previously accepted policies on the same complete set of development seeds. For each historical policy, the candidate must achieve a strictly higher mean development score and a higher task score on at least half of the development seeds.

\textbf{Population: Population search from a shared parent.} In each generation, two isolated candidates are generated from the same parent policy. After completing the official development evaluation, the candidate with the higher raw development score is selected. At most one candidate advances per generation.

\textbf{Force: Forced continuation with context retention.} The agent is forced to run for 30 rounds. If it declares completion early, the controller always injects a requirement to continue learning, while retaining the same session context throughout.

\paragraph{Advanced Strategies}
In addition to these four alternative construction paths, we investigate two execution strategies on the current meta-harness that increase model performance by allocating more training effort.

\textbf{Reset: Continuation with a fresh context.} After each run, the policy at the end of that run is used as the initial policy for the next run. Only the current best policy and development evaluation records are carried over; the original conversational context is not retained.

\textbf{Swarm: Parallel repeated learning.} Five runs are launched in parallel using the same learning protocol. For each task, the result with the highest raw development score is selected, and we report the results obtained by selecting the best across all five runs.

\begin{table}[t]
\centering
{\small
\caption{Performance and cost of the four alternative learning-control paths (Meta, History, Population, and Force) and the Standard, Reset, and Swarm PDMetaHarness strategies. Scores are test-evaluation results after development-set selection for each task; costs include the initial execution of each alternative and all additional reruns following failures.}\label{tab:learning-control-paths}
\par\smallskip
\setlength{\tabcolsep}{2pt}
\begin{tabular}{@{}
  >{\raggedright\arraybackslash}p{(\linewidth - 10\tabcolsep) * \real{0.28}}
  >{\raggedleft\arraybackslash}p{(\linewidth - 10\tabcolsep) * \real{0.145}}
  >{\raggedleft\arraybackslash}p{(\linewidth - 10\tabcolsep) * \real{0.145}}
  >{\raggedleft\arraybackslash}p{(\linewidth - 10\tabcolsep) * \real{0.145}}
  >{\raggedleft\arraybackslash}p{(\linewidth - 10\tabcolsep) * \real{0.145}}
  >{\raggedleft\arraybackslash}p{(\linewidth - 10\tabcolsep) * \real{0.14}}@{}}
\toprule
\begin{minipage}[b]{\linewidth}\raggedright
Method
\end{minipage} & \begin{minipage}[b]{\linewidth}\raggedleft
Contin. control (16)
\end{minipage} & \begin{minipage}[b]{\linewidth}\raggedleft
Discrete control (11)
\end{minipage} & \begin{minipage}[b]{\linewidth}\raggedleft
Overall (27)
\end{minipage} & \begin{minipage}[b]{\linewidth}\raggedleft
Build time (h)
\end{minipage} & \begin{minipage}[b]{\linewidth}\raggedleft
Infer\-ence time (s)
\end{minipage} \\
\midrule
\textbf{Alternatives} & & & & & \\
Meta & 0.501 & 0.852 & 0.644 & 236.3 & 95 \\
History & 0.529 & 0.836 & 0.654 & 207.2 & 112 \\
Population & 0.569 & 0.866 & 0.690 & 176.5 & 124 \\
Force & 0.596 & 0.872 & 0.709 & 121.9 & 1567 \\
\textbf{PDMetaHarness} & & & & & \\
Standard & 0.618 & 0.864 & 0.718 & 38.0 & 110 \\
Reset & \underline{0.644} & \textbf{0.889} & \underline{0.744} & 172.0 & 133 \\
Swarm & \textbf{0.652} & \underline{0.882} & \textbf{0.746} & 400.3 & 127 \\
\bottomrule
\end{tabular}}
\end{table}

\paragraph{Results}
The table summarizes the learning control paths and the results of development-set selection. The four alternatives achieve overall scores between 0.644 and 0.709, none surpassing our original PDMetaHarness. With more extensive training, Reset and Swarm achieve overall scores of 0.744 and 0.746, respectively, despite incurring higher training costs. These results represent improvements of approximately 3.6\% and 3.9\% over the standard method. The results show that more extensive training can further improve performance. In practice, Reset may offer a favorable trade-off between performance gains and training cost.

\subsection{\emph{RQ4}: Can the Capability of Teacher \(T_\theta\) Be Distinguished from That of Optimizer \(\mathcal O_\theta\)?}\label{planner-replacement-and-knowledge-distillation}

To investigate the role of the distillation teacher, we design a distinctive isolation scheme that separates the effects of different models and then uses multiple models in combination to solve each problem. Under the standard setting, GLM-5.3 is fixed as the optimizer, while only the teacher used to distill candidate programs is replaced. This setup separates the model that proposes programs from the model that conducts experiments, allowing the capability of a model as a distillation teacher to be analyzed independently.

\begin{table}[t]
\centering
{\small
\caption{Test-evaluation performance when the GLM-5.3 tool agent is fixed as the optimizer and only the distillation teacher is replaced. The continuous and discrete columns are grouped by action type, so that corresponding settings can be compared directly across teachers. Failed tasks still enter test evaluation through their frozen submissions rather than being discarded, and the overall scores are computed over all 27 tasks.}\label{tab:teacher-replacement}
\par\smallskip
\setlength{\tabcolsep}{2pt}
\begin{tabular}{@{}
  >{\raggedright\arraybackslash}p{(\linewidth - 10\tabcolsep) * \real{0.34}}
  >{\raggedleft\arraybackslash}p{(\linewidth - 10\tabcolsep) * \real{0.135}}
  >{\raggedleft\arraybackslash}p{(\linewidth - 10\tabcolsep) * \real{0.135}}
  >{\raggedleft\arraybackslash}p{(\linewidth - 10\tabcolsep) * \real{0.135}}
  >{\raggedleft\arraybackslash}p{(\linewidth - 10\tabcolsep) * \real{0.135}}
  >{\raggedleft\arraybackslash}p{(\linewidth - 10\tabcolsep) * \real{0.12}}@{}}
\toprule
\begin{minipage}[b]{\linewidth}\raggedright
Model
\end{minipage} & \begin{minipage}[b]{\linewidth}\raggedleft
Contin. control (16)
\end{minipage} & \begin{minipage}[b]{\linewidth}\raggedleft
Discrete control (11)
\end{minipage} & \begin{minipage}[b]{\linewidth}\raggedleft
Overall (27)
\end{minipage} & \begin{minipage}[b]{\linewidth}\raggedleft
Build time (h)
\end{minipage} & \begin{minipage}[b]{\linewidth}\raggedleft
Infer\-ence time (s)
\end{minipage} \\
\midrule
GLM-5.3 (single) & \textbf{0.618} & \underline{0.864} & \textbf{0.718} & 38.0 & 110 \\
GLM-5.3 + DeepSeek-V4.1-Flash & \underline{0.594} & \textbf{0.865} & \underline{0.705} & 49.9 & 101 \\
GLM-5.3 + Kimi-K2.6 & 0.501 & 0.848 & 0.642 & 43.8 & 116 \\
GLM-5.3 + Qwen3.6-35B-A3B & 0.550 & 0.785 & 0.646 & 25.3 & 134 \\
\bottomrule
\end{tabular}}
\end{table}

Table~\ref{tab:teacher-replacement} shows a clear positive relationship between the capability of the distillation teacher and the final evaluation performance. Measured by the standalone PD total score in Table~\ref{tab:main-results}, GLM-5.3 and DeepSeek-V4.1-Flash achieve relatively strong overall performance, whereas Kimi-K2.6 and Qwen3.6-35B-A3B perform relatively poorly overall, indicating a substantial generational gap between the two groups of models.

Although this gap is reduced after replacing the optimizer with GLM-5.3, it remains clearly identifiable overall. The two models in the higher-performing group achieve scores of 0.718 and 0.705, while those in the lower-performing group achieve 0.642 and 0.646. These results are consistent with the expectation that the capabilities of the LLM teacher are distilled into the student program: stronger distillation teachers provide more useful task priors, control structures, and parameter candidates. After passing through the same filtering process, the resulting student programs also achieve better evaluation results on the test set. This shows that a pure optimizer view cannot fully explain teacher capability, whereas isolating the teacher role can.

\subsection{\emph{RQ5}: How Does the Learning Process Justify the Distillation View?}\label{learning-process}

\paragraph{Capability Acquisition by Models Is Breakthrough-Driven.}\label{breakthrough-driven-capability-acquisition}

Table~\ref{tab:main-results} reports the execution results for 27 tasks across 10 models. After the initial policy, these runs made a total of 3,643 valid attempts, and 244 runs achieved positive gains. We ask whether models acquire capabilities through breakthroughs or gradual improvement.

\begin{figure}
\centering
\pandocbounded{\includegraphics[keepaspectratio]{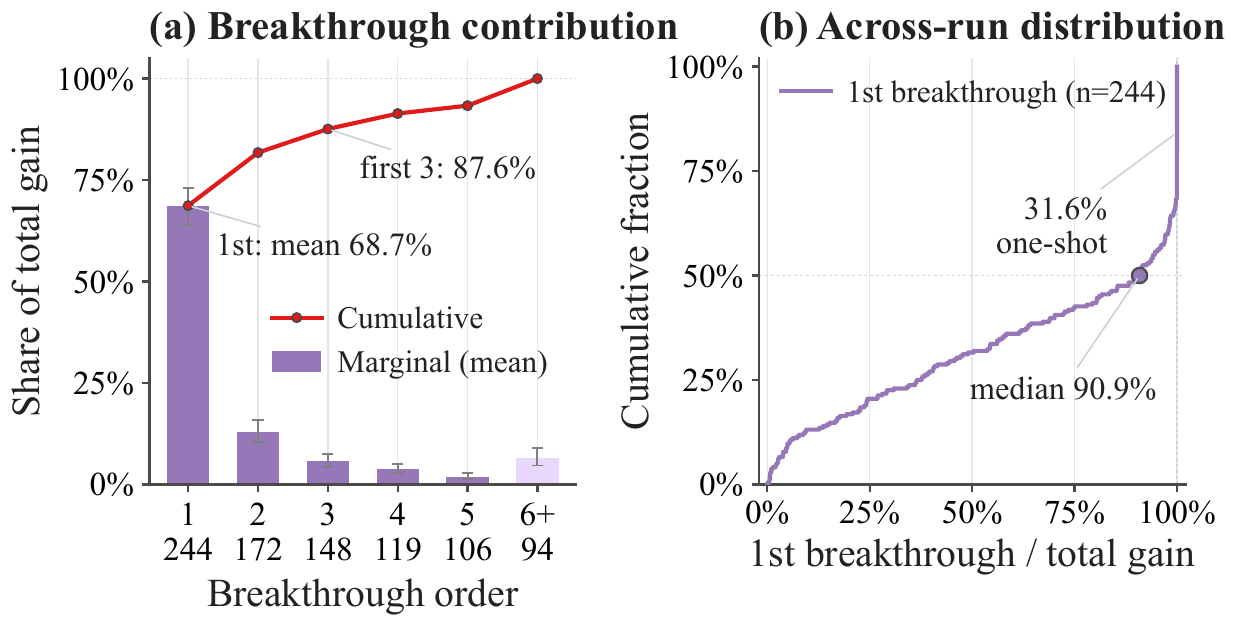}}
\caption{Contribution of successive breakthroughs. (a) Mean share of total development gain by breakthrough order, with 95\% bootstrap intervals and the cumulative share; (b) Distribution of the first-breakthrough share across 244 positive-gain runs.}\label{fig:breakthrough-contribution}
\end{figure}

To answer this question, Figure~\ref{fig:breakthrough-contribution}(a) presents the main result: the first improvement in policy performance accounts for an average of 68.7\% of the total gain over a run, while the first three improvements account for an average of 87.6\%. The across-run distribution in Figure~\ref{fig:breakthrough-contribution}(b) reveals an even more striking pattern. Based on the median, the first improvement alone accounts for 90.9\% of a run's total gain. Moreover, 31.6\% of the runs achieve 100\% of their total gain in a single attempt, indicating that nearly all gains in many runs come from this one event. These results show that learning does not typically proceed through steady, incremental improvements along a single trajectory. Instead, it is usually characterized by a sharp breakthrough, with gains concentrated almost entirely in the model's first policy improvement. Models almost always improve through sudden insight rather than gradual accumulation, so capability acquisition is more one-shot than repeated tuning.

\paragraph{Structural Modifications Are Low-Hit but High-Reward.}\label{structural-edits-low-hit-high-reward}

We divide policy modifications into two categories, structural modifications and parameter modifications, and examine them using this binary classification. Specifically, we parse the policy program files before and after each modification and compare their code skeletons, ignoring numerical values. If the skeletons are identical, we classify the modification as a parameter modification; otherwise, we classify it as a structural modification. Structural modifications account for 52.0\% of all modifications, while parameter modifications account for 48.0\%. We ask how structural and parameter modifications to policies affect model capabilities.

\begin{figure}
\centering
\pandocbounded{\includegraphics[keepaspectratio]{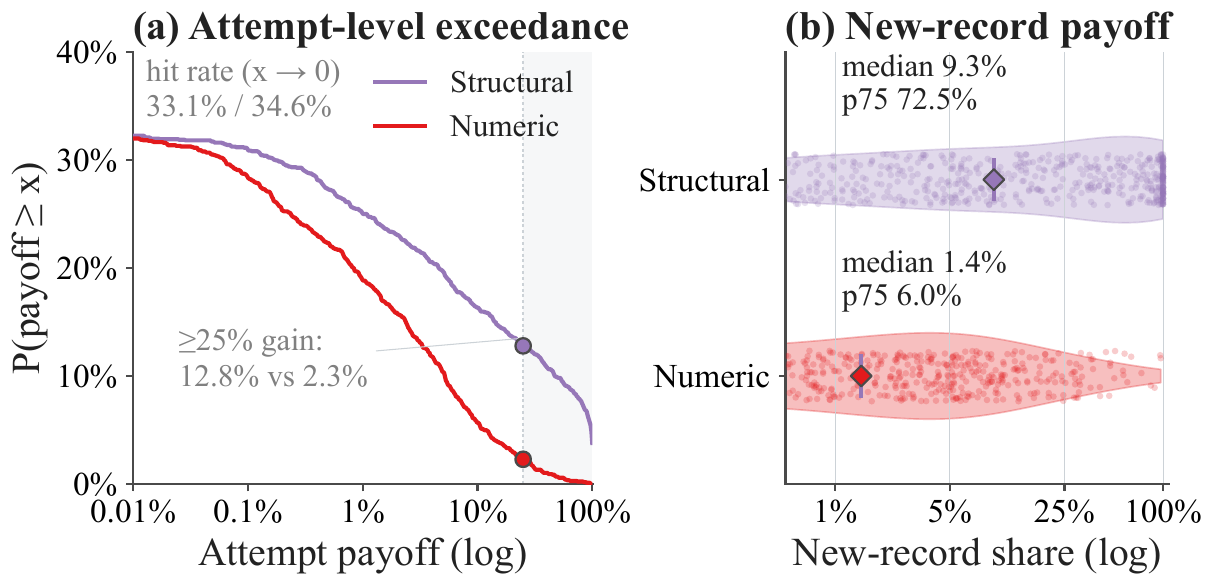}}
\caption{Structural modifications as low-hit, high-reward events. (a) Attempt-level exceedance, the probability that an attempt's payoff reaches a given share of run gain; structural and parameter modifications have similar success rates (33.1\% versus 34.6\%). (b) Distribution of event share conditional on setting a new record; the median contribution is 9.28\% for structural modifications versus 1.44\% for parameter modifications, and the 75th-percentile shares are 72.5\% and 6.0\%. Structural modifications succeed slightly less often but have a substantially heavier payoff tail.}\label{fig:structural-lottery}
\end{figure}

\begin{figure}
\centering
\pandocbounded{\includegraphics[keepaspectratio]{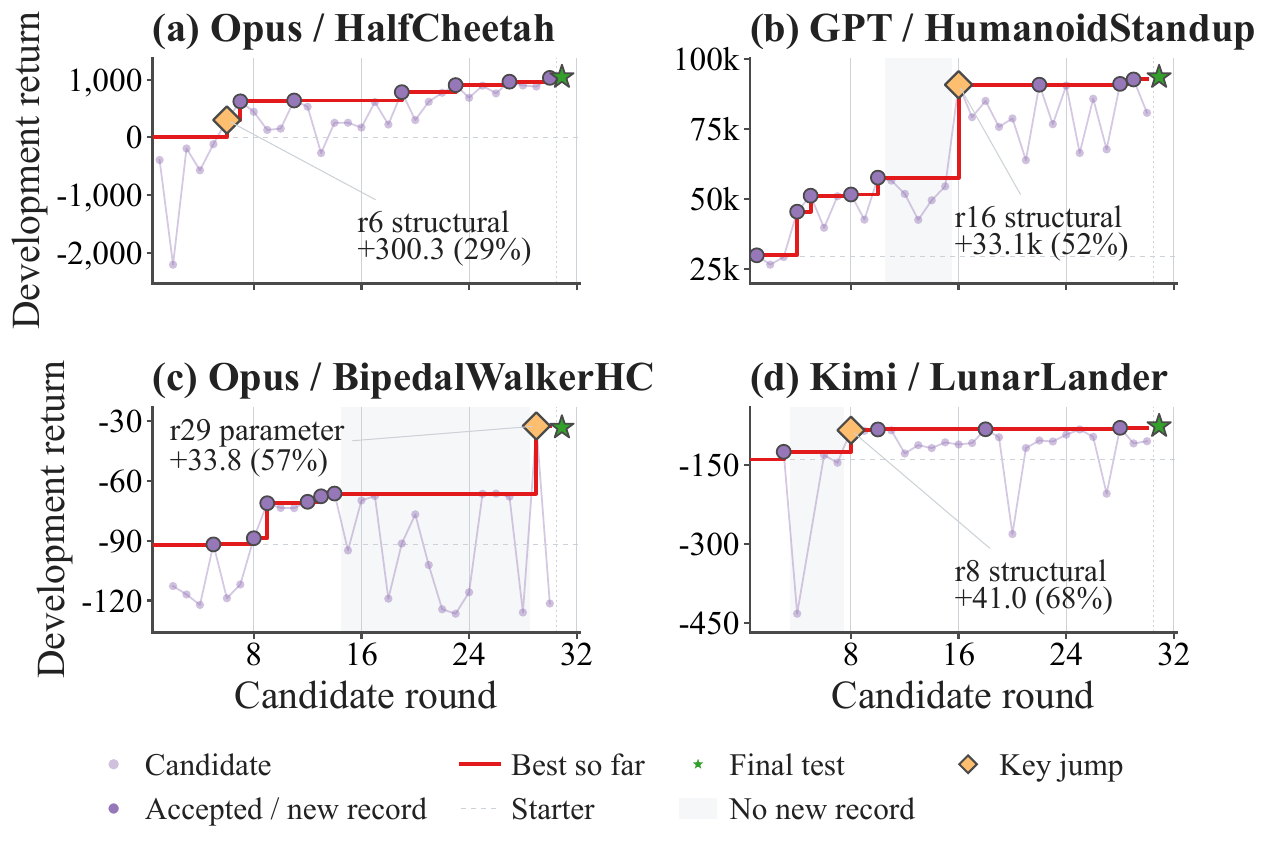}}
\caption{Four representative learning trajectories: an early structural jump, a later structural jump, a parameter jump near the end of the budget, and a case with substantial gain but weak test evaluation. Gray marks candidate development scores, blue points mark accepted candidates that set records, the orange line marks the historical best, and the dashed line marks the initial policy. Gray shading indicates the pre-jump interval without a new record.}\label{fig:trajectory-examples}
\end{figure}

As shown in Figure~\ref{fig:structural-lottery}(a), structural and parameter modifications have similar overall success rates: 33.1\% of structural modification attempts improve on the previous rule, compared with 34.6\% of parameter modification attempts. Structural modifications therefore have a slightly lower success rate. Their advantage lies instead in the magnitude of their effects. Structural and parameter modifications produce gains of at least 25\% of the total gain in 12.8\% and 2.3\% of attempts, respectively. Further, Figure~\ref{fig:structural-lottery}(b) shows that the median fraction of total improvement contributed by a successful modification is 9.28\% for structural modifications, compared with 1.44\% for parameter modifications. At the 75th percentile, the corresponding values are 72.5\% and 6.0\%. These results indicate that structural modifications account for most of the development gains. Structural modifications thus exhibit a low-hit, high-reward pattern: they succeed slightly less often, but their gains have a substantially heavier tail. Parameter modifications, by contrast, tend to provide small, somewhat more frequent adjustments. As an illustrative example, Figure~\ref{fig:trajectory-examples} presents four representative learning trajectories. Overall, performance comes more from structural adjustments than from parameter fine-tuning, and teacher knowledge and capability largely determine these structural outcomes.

\subsection{\emph{RQ6}: How Does the Distillation View Enable Exploration Beyond Pure Performance Optimization?}\label{stylized-learning-and-an-interpretable-policy-pool}

\paragraph{Style Objectives.}\label{style-objectives}

We propose a style-distillation training method that exploits the LLM teacher's ability to interpret behavioral characteristics. During distillation, the model is instructed to follow a particular behavioral style. This approach can support the generation of personalized characters or levels in domains such as games.

To illustrate how task performance and behavioral style jointly shape policy construction, and how the resulting programs can be used to generate a population of players with diverse styles, we use Pong, a classic adversarial game in PDBench, as an example.

We refer to longer and shorter scoring rallies as \emph{conservative} and \emph{adventurous}, respectively, and to lower and higher paddle reversal tendencies as \emph{stable} and \emph{change-oriented}. These labels correspond to measurable model behaviors and align closely with intuitive human perceptions of these styles.

We characterize Pong style using two behavioral proxies that can be recomputed from trajectories: rally length and the frequency of actual paddle reversals. Let \(N_{\mathrm{vis}}\), \(N_{\mathrm{rally}}\), and \(N_{\mathrm{rev}}\) denote the number of ball-visible decision steps, completed scoring rallies, and actual paddle reversals, respectively, over the full development match. We define
\begin{equation}
L({S_\phi})=\frac{N_{\mathrm{vis}}}{N_{\mathrm{rally}}},
\qquad
V({S_\phi})=1000\frac{N_{\mathrm{rev}}}{N_{\mathrm{vis}}}.
\end{equation}
The \emph{conservative/adventurous} metric \(L({S_\phi})\) measures the number of visible decision steps per scoring rally, while the \emph{stable/change-oriented} metric \(V({S_\phi})\) measures the number of actual paddle reversals per thousand visible decision steps.

\paragraph{Training and Capability Ranking.}\label{training-and-capability-ranking}

We combine the two style dimensions to form four training-objective quadrants and launch three training runs for each quadrant, for a total of 12 runs. Before each run, the target style is specified to the model through task instructions and skill descriptions. During training, development feedback includes both the score difference and the style metrics.

For each new candidate, the style objective is incorporated into learning through a style-specific prompt. This effectively adds a distinct form of teacher guidance, distilling the teacher's understanding of a particular policy style. It also introduces a policy-style objective alongside the objective of achieving the highest score. We expect that, if an optimal style exists, the model can learn both objectives simultaneously.

Retaining only the single highest-scoring program would discard behavioral diversity among intermediate candidates. We therefore collect all policies generated during development as evaluation candidates, forming a development style calibration set of 198 policies. We assign each policy to one of four quadrants according to the strength of its style characteristics. We then evaluate every pair of code-based policies through direct matches in Pong and use the outcomes to rank their capabilities. In total, we conduct 2,000 matches and rank the policies using the Elo rating system.

\paragraph{Learning Results.}\label{style-results}

A key advantage of program distillation is that it allows us to investigate whether an optimal control style exists for a given class of tasks.

\begin{figure}
\centering
\pandocbounded{\includegraphics[keepaspectratio]{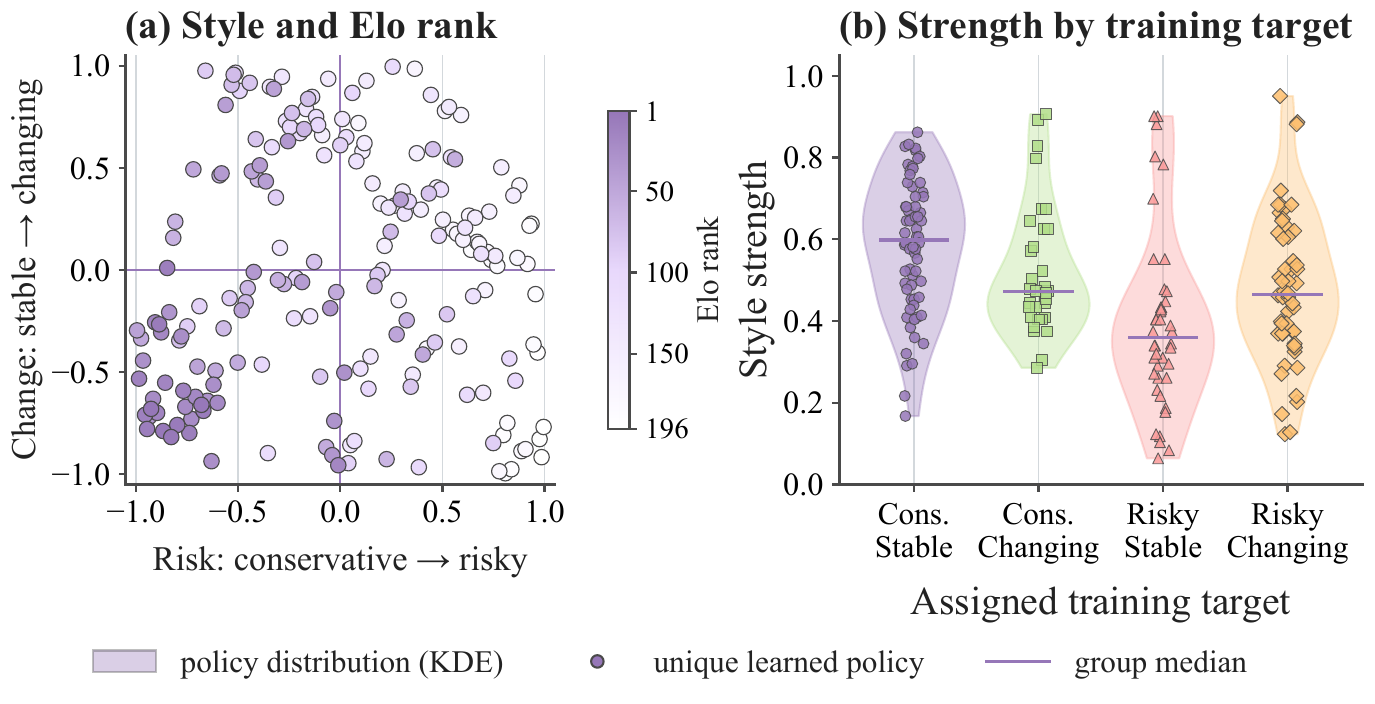}}
\caption{Measured style, internal Elo rank, and style strength, reported by assigned training target. (a) Measured style positions, where darker color indicates a higher internal Elo rank. (b) Style strength under each assigned target; across all training methods, the conservative-stable style is maintained more consistently than the other styles, indicating that this target is easier to preserve during training.}\label{fig:style-elo}
\end{figure}

Figure~\ref{fig:style-elo}(a) shows a clear pattern in model capability: nearly all policies with the highest Elo ratings fall in the conservative-stable quadrant. Moreover, the more strongly a policy exhibits this style, the higher its capability. Figure~\ref{fig:style-elo}(b) shows the degree of alignment between model behavior and the target style during training. Because distillation aims to align the model with the teacher while improving its performance, a natural conflict between the style objective and high performance would suggest that achieving high performance while maintaining that style may be infeasible. We find that the conservative-stable style is maintained more consistently than the other styles across all training methods. This result suggests an inherent conflict between the two objectives that training may not be able to resolve simultaneously. Taken together, these results indicate that the conservative-stable style is not only beneficial for improving capability, but may also represent the optimal style. This demonstrates that style distillation enables stylistic learning and identifies the optimal style.

\section{Related Work}

\label{sec:related-work}

\subsection{Knowledge Distillation}

Knowledge distillation transfers capabilities from a computationally expensive teacher to a more efficient student \cite{hinton2015distilling}. In sequential decision making, policy distillation compresses a teacher policy or consolidates multiple expert policies into a single student \cite{rusu2015policy}. System 2 distillation similarly transfers explicit reasoning capabilities to direct-answer models, thereby reducing inference-time computation \cite{yu2024system2}. Related studies distill neural policies into symbolic programs. PIRL searches for interpretable policies in a domain-specific language, guided by a neural policy \cite{verma2018pirl}, while VIPER extracts verifiable decision trees from a teacher policy and its value function \cite{bastani2018viper}. Code as Policies further shows that LLMs can generate executable programs that invoke perception and control interfaces \cite{liang2023code}. Our work extends these ideas to interactive policy construction from System 2 to System 0. The LLM supplies task knowledge and proposes or revises policy code, while the resulting program preserves the acquired behavior and executes independently at deployment.

\subsection{Feedback-Driven Program Evolution}

Feedback-driven methods improve behavior or executable artifacts without updating model parameters. Self-Refine revises outputs using model-generated feedback, whereas Reflexion records linguistic reflections from previous attempts to guide subsequent behavior \cite{madaan2023selfrefine,shinn2023reflexion}. FunSearch and AlphaEvolve combine LLM-based generation with automated evaluation to search over programs \cite{romeraparedes2024funsearch,novikov2025alphaevolve}. EoH evolves program descriptions and implementations, while ReEvo exploits performance differences among candidates to guide further search \cite{liu2024eoh,ye2024reevo}. Several studies apply this paradigm to policy code. LangProp uses execution errors and task feedback to optimize policies \cite{ishida2024langprop}, and Generative Code Optimization revises Atari policies based on trajectories and linguistic feedback \cite{kuang2025generative}. PromptPO generates and selects Python policies from trajectory feedback, achieving strong sample efficiency across multiple tasks \cite{hatgiskessell2026promptpo}. These studies highlight the importance of incorporating both the strategies being optimized and the full environmental context into the feedback process, an idea closely related to the concept of a harness. However, no standardized benchmark currently enables fair comparisons across different approaches. Consequently, it remains difficult to disentangle the contributions of the underlying model from those of the harness architecture. This limitation motivates the need for a unified, model- and harness-agnostic meta-system that organizes the entire harness-learning process systematically. Addressing this challenge is a central objective of our work.

\subsection{Efficient Decision Making and Control}

Real-time control requires balancing action quality against inference latency. Real-time reinforcement learning explicitly models the concurrent evolution of agent computation and the environment, exposing limitations of simulators that wait for each action before proceeding \cite{ramstedt2019real}. Studies of LLM-based decision making likewise report a trade-off between response speed and decision quality \cite{kang2025speed,chen2026system1arena}. Hierarchical systems mitigate this trade-off by assigning high-level reasoning to LLMs and low-level execution to faster controllers or skills \cite{ahn2022saycan,pubgally2026}. Action chunking further reduces latency by overlapping action execution with generation \cite{black2025realtime}. An alternative approach is to compile model capabilities into independently executable programs. Cardiverse optimizes LLM-generated programs to reduce repeated model calls during gameplay \cite{li2025cardiverse}. Code to Control synthesizes reactive controllers and optimizes their parameters without requiring LLM calls during execution \cite{ahmed2026codetocontrol}. Related program-search methods have also been applied to active flow control and traffic signal control \cite{garnier2026flow,wang2026traffic}. In contrast to the prevailing view that action quality and inference latency are inherently coupled, we theoretically show that this trade-off is not fundamental. By distilling capabilities from System 2 to System 0, our approach produces policies that improve decision quality while reducing inference latency.

\section{Conclusion}\label{conclusion}

We propose Program Distillation (PD), a knowledge distillation framework that converts LLM-guided reasoning into standalone decision programs. PD combines a frozen LLM with iterative program refinement and supports reproducible evaluation through PDBench and PDMetaHarness. Across 27 control tasks, PD achieves a normalized mean score of 0.728, surpassing PPO and online-decision baselines, while producing CPU-executable programs with no model calls at deployment. Further analyses show that stronger teachers yield better programs, early breakthroughs account for most performance gains, and style distillation enables interpretable behavioral control. These results establish PD as an effective and deployable approach for transforming model-guided experimentation into reusable decision policies.

\bibliographystyle{ACM-Reference-Format}
\bibliography{ref}

\appendix

\section{Task Protocol}\label{complete-task-protocol}

The experimental environment is fixed to Gymnasium 1.3.0, MuJoCo 3.3.7, Box2D 2.3.10, ALE 0.12.0, and NumPy 2.4.6. Digests of the MuJoCo XML files are stored with the protocol. Table~\ref{tab:task-protocol} lists the environment ID, action space, decision limit, and scoring anchors for each of the 27 tasks.
LunarLander disables wind, standard BipedalWalker disables hardcore terrain, Taxi disables rain and passenger variation, and FrozenLake retains stochastic slipping. Pong uses frame skip 4 and sticky-action probability 0.250.

\begin{table*}[t]
\centering
\caption{Complete task protocol. Each task is listed with its environment ID, action space, maximum decision steps, and lower/upper scoring anchors. The \texttt{gym\_} prefix is omitted from task names.}\label{tab:task-protocol}
\par\smallskip
{\footnotesize
\begin{tabular}{@{}
  >{\raggedright\arraybackslash}p{(\linewidth - 8\tabcolsep) * \real{0.23}}
  >{\raggedright\arraybackslash}p{(\linewidth - 8\tabcolsep) * \real{0.24}}
  >{\raggedright\arraybackslash}p{(\linewidth - 8\tabcolsep) * \real{0.19}}
  >{\raggedleft\arraybackslash}p{(\linewidth - 8\tabcolsep) * \real{0.13}}
  >{\raggedright\arraybackslash}p{(\linewidth - 8\tabcolsep) * \real{0.21}}@{}}
\toprule
\begin{minipage}[b]{\linewidth}\raggedright
Task
\end{minipage} & \begin{minipage}[b]{\linewidth}\raggedright
Environment ID
\end{minipage} & \begin{minipage}[b]{\linewidth}\raggedright
Action space
\end{minipage} & \begin{minipage}[b]{\linewidth}\raggedleft
Maximum decision steps
\end{minipage} & \begin{minipage}[b]{\linewidth}\raggedright
Lower / upper anchor
\end{minipage} \\
\midrule
atari\_pong & ALE/Pong-v5 & 6 discrete actions & 100,000; ends at 11 points & -11.000 / 11.000 \\
cartpole & CartPole-v1 & 2 discrete actions & 500 & 0.000 / 500.000 \\
mountain\_car & MountainCar-v0 & 3 discrete actions & 200 & -200.000 / -80.000 \\
mountain\_car\_continuous & MountainCarContinuous-v0 & 1 dimension, range {[}-1, 1{]} & 999 & -100.000 / 100.000 \\
pendulum & Pendulum-v1 & 1 dimension, range {[}-2, 2{]} & 200 & -1600.000 / 0.000 \\
acrobot & Acrobot-v1 & 3 discrete actions & 500 & -500.000 / -80.000 \\
lunar\_lander & LunarLander-v3 & 4 discrete actions & 1,000 & -300.000 / 300.000 \\
bipedal\_walker & BipedalWalker-v3 & 4 dimensions, range {[}-1, 1{]} & 1,600 & -150.000 / 300.000 \\
taxi & Taxi-v4 & 6 discrete actions & 200 & -200.000 / 20.000 \\
frozen\_lake & FrozenLake8x8-v1 & 4 discrete actions & 200 & 0.000 / 1.000 \\
blackjack & Blackjack-v1 & 2 discrete actions & 32 & -1.000 / 1.000 \\
cliff\_walking & CliffWalking-v1 & 4 discrete actions & 200 & -20000.000 / -13.000 \\
cliff\_walking\_slippery & CliffWalkingSlippery-v1 & 4 discrete actions & 200 & -20000.000 / -13.000 \\
frozen\_lake4x4 & FrozenLake-v1 & 4 discrete actions & 100 & 0.000 / 1.000 \\
lunar\_lander\_continuous & LunarLanderContinuous-v3 & 2 dimensions, range {[}-1, 1{]} & 1,000 & -300.000 / 300.000 \\
bipedal\_walker\_hardcore & BipedalWalkerHardcore-v3 & 4 dimensions, range {[}-1, 1{]} & 2,000 & -150.000 / 300.000 \\
inverted\_pendulum & InvertedPendulum-v5 & 1 dimension, range {[}-3, 3{]} & 1,000 & 0.000 / 1000.000 \\
inverted\_double\_pendulum & InvertedDoublePendulum-v5 & 1 dimension, range {[}-1, 1{]} & 1,000 & 0.000 / 10000.000 \\
reacher & Reacher-v5 & 2 dimensions, range {[}-1, 1{]} & 50 & -100.000 / 0.000 \\
pusher & Pusher-v5 & 7 dimensions, range {[}-2, 2{]} & 100 & -200.000 / 0.000 \\
swimmer & Swimmer-v5 & 2 dimensions, range {[}-1, 1{]} & 1,000 & -100.000 / 400.000 \\
hopper & Hopper-v5 & 3 dimensions, range {[}-1, 1{]} & 1,000 & -100.000 / 4000.000 \\
half\_cheetah & HalfCheetah-v5 & 6 dimensions, range {[}-1, 1{]} & 1,000 & -1000.000 / 12000.000 \\
walker2d & Walker2d-v5 & 6 dimensions, range {[}-1, 1{]} & 1,000 & -100.000 / 5000.000 \\
ant & Ant-v5 & 8 dimensions, range {[}-1, 1{]} & 1,000 & -1000.000 / 6000.000 \\
humanoid & Humanoid-v5 & 17 dimensions, range {[}-0.400, 0.400{]} & 1,000 & 0.000 / 6000.000 \\
humanoid\_standup & HumanoidStandup-v5 & 17 dimensions, range {[}-0.400, 0.400{]} & 1,000 & 0.000 / 150000.000 \\
\bottomrule
\end{tabular}
}
\end{table*}

The policy interface prohibits external-module imports, classes, nested helper functions, asynchronous calls, mutable global state, file or network access, pretrained weights, and seed-indexed lookup tables. Gymnasium policies may use only restricted built-ins and approved mathematical operations. Discrete actions must belong to the legal action set. Continuous actions must have the correct dimensionality, contain finite values, and lie within the specified range.

\section{Cumulative Gain over Observed Attempts}\label{cumulative-gain-appendix}

\begin{figure}
\centering
\pandocbounded{\includegraphics[keepaspectratio]{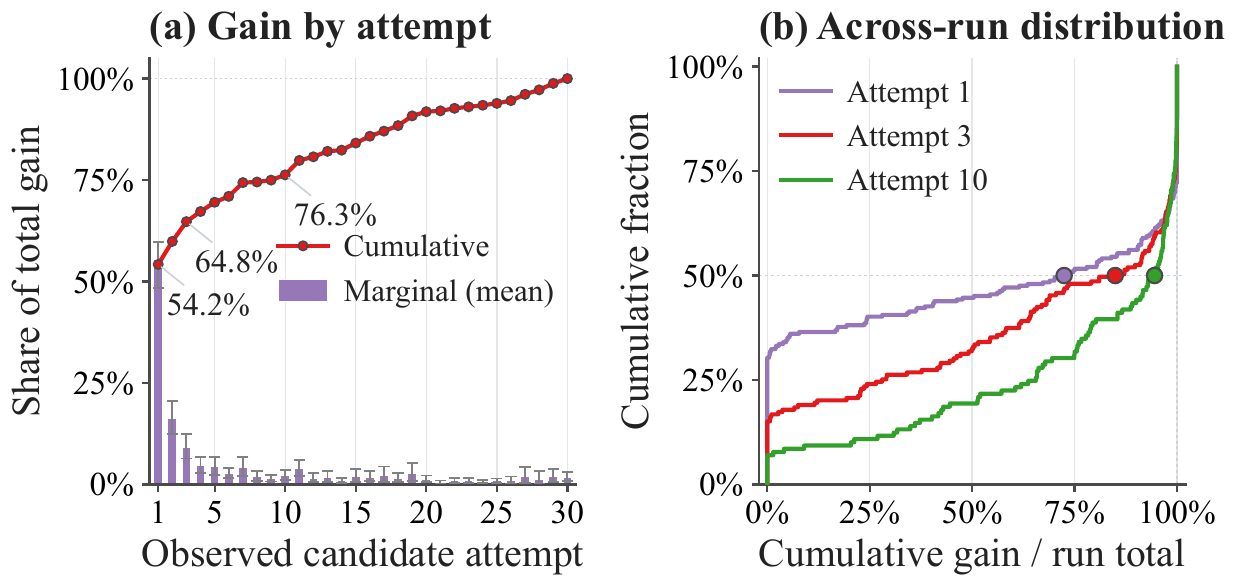}}
\caption{Cumulative gain over observed attempts. (a) Mean marginal increment and cumulative share of total development gain by observed candidate attempt. (b) Distribution of cumulative gain at attempts 1, 3, and 10.}\label{fig:cumulative-gain}
\end{figure}

Figure~\ref{fig:cumulative-gain} reports cumulative gain over observed attempts. Panel (a) shows the mean marginal increment and the cumulative share of total development gain by observed candidate attempt, while panel (b) shows the distribution of cumulative gain at attempts 1, 3, and 10. Gains are front-loaded relative to both the candidate budget and the observed trajectory: models obtain 72.7\%--97.4\% of their gains within the first 25.0\% of the budget, but only 10.1\%--64.9\% within the first 25.0\% of the observed trajectory. Early termination directly affects this difference, because a run with a 30-round budget but only one executed round places its sole experiment at 3.3\% of the budget and 100.0\% of the observed trajectory.

\section{The Echo Court Local Game}\label{the-echo-court-local-game-and-program-deployment}

\subsection{From the Policy Pool to a Desktop Game}\label{f.1-from-the-policy-pool-to-a-desktop-game}

Style distillation enables a further use of the resulting policy pool: we design a desktop game around it. The game contains 500 virtual players, each with a distinct policy and a complete Elo ranking. Playing the game locally reproduces the experience of online matchmaking.

Echo Court is a local Pong application implemented in Python and pygame. It supports human-versus-policy and two-player games, spectating, rankings, matchmaking, and persistent match records. The application packages source files and JSON definitions for 500 composite players, 197 base policies, and a frozen composition runtime. We combine different composition mechanisms to generate this larger set of 500 players.

\subsection{Local Rankings and Matchmaking}\label{f.3-local-rankings-and-matchmaking}

\begin{figure}
\centering
\pandocbounded{\includegraphics[keepaspectratio]{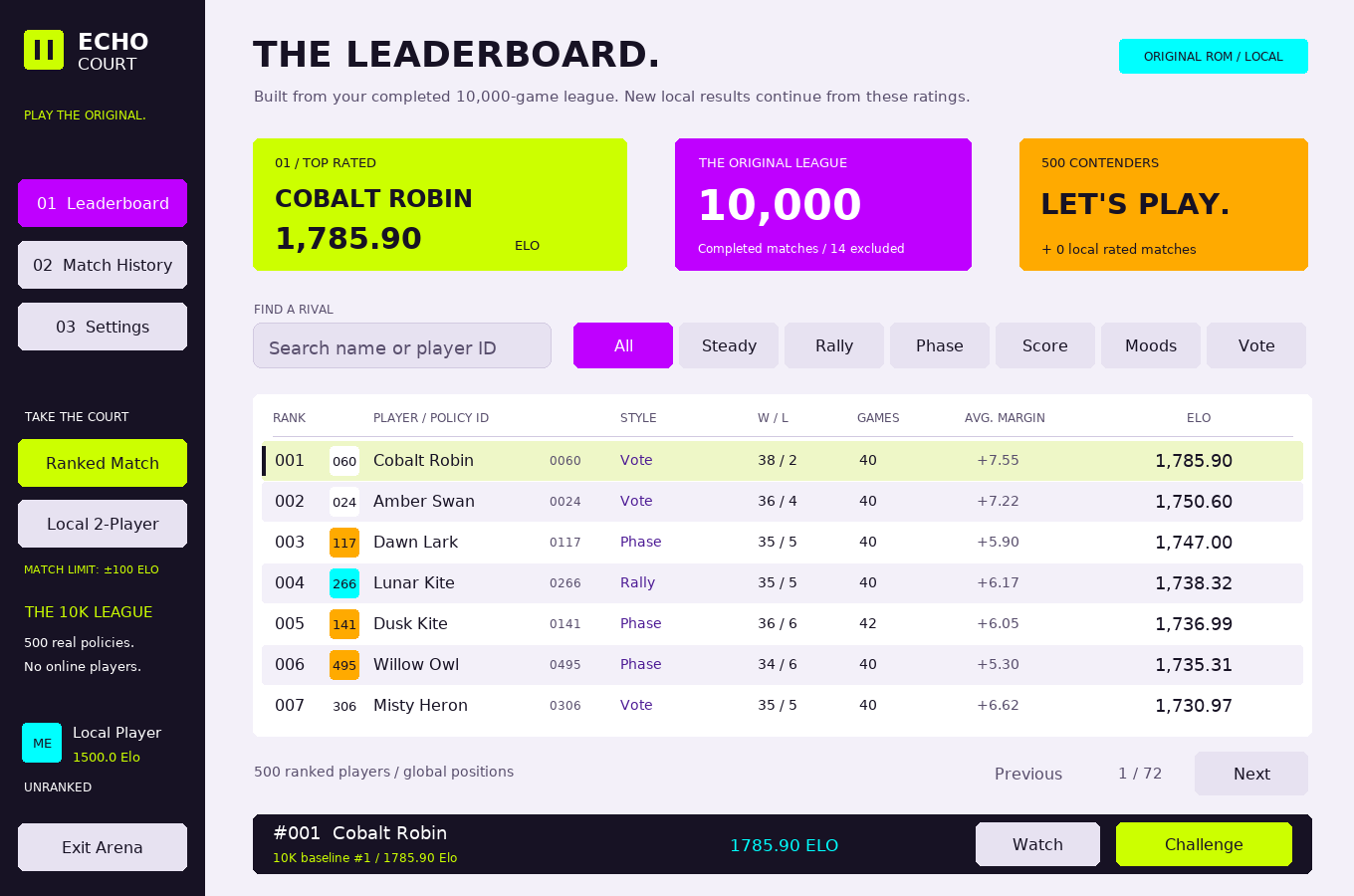}}
\caption{Complete leaderboard for a snapshot of the 500-player Echo Court league. Players are ordered by Elo, with wins, losses, and mean score difference per game retained.}\label{fig:leaderboard}
\end{figure}

In a new season, all 500 policy players and new users start at an Elo rating of 100.000. Automatic matchmaking first samples uniformly from the complete candidate set within a rating difference of 50.000. If this set is empty, the range expands to 100.000; if it remains empty, matchmaking stops. When a match is confirmed, both ratings are checked again. Upon completion, ratings are updated using zero-sum Elo with update factor 32 and scale 400. Match records preserve the matchmaking interval, candidate count, both pre-match ratings, and provenance. Figure~\ref{fig:leaderboard} shows the complete leaderboard for a snapshot.

\subsection{A 50-Match Experience}\label{f.4-a-50-match-fixed-policy-experience}

We use a virtual player to demonstrate the full game flow. Rally100 is a virtual robot player that uses the unoptimized initial program from the reference commit to control the right paddle. It completes 50 matches through the actual interface flow of entry, matchmaking, confirmation, play, and settlement.

Across the 50 valid matches, Rally100 faced 42 distinct opponents, won 24 and lost 26, and accumulated a score of 430:427 for a mean score difference of +0.060 per match. Its Elo rating moved from 100.000 to 63.998.

\begin{figure}
\centering
\pandocbounded{\includegraphics[keepaspectratio]{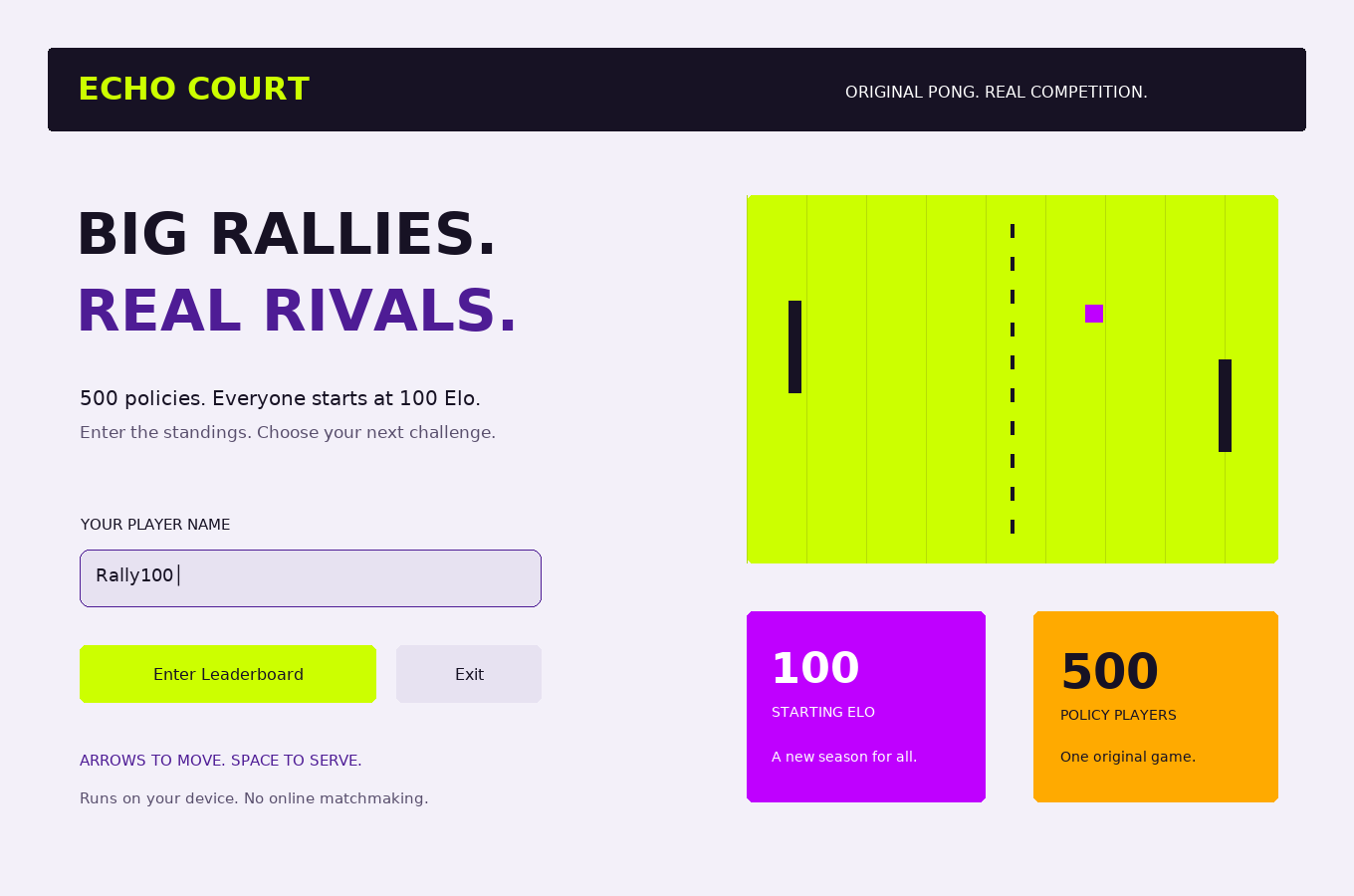}}
\caption{A new player entering the local game; screenshot produced by off-screen rendering of the actual pygame window.}\label{fig:echo-court-entry}
\end{figure}

\begin{figure}
\centering
\pandocbounded{\includegraphics[keepaspectratio]{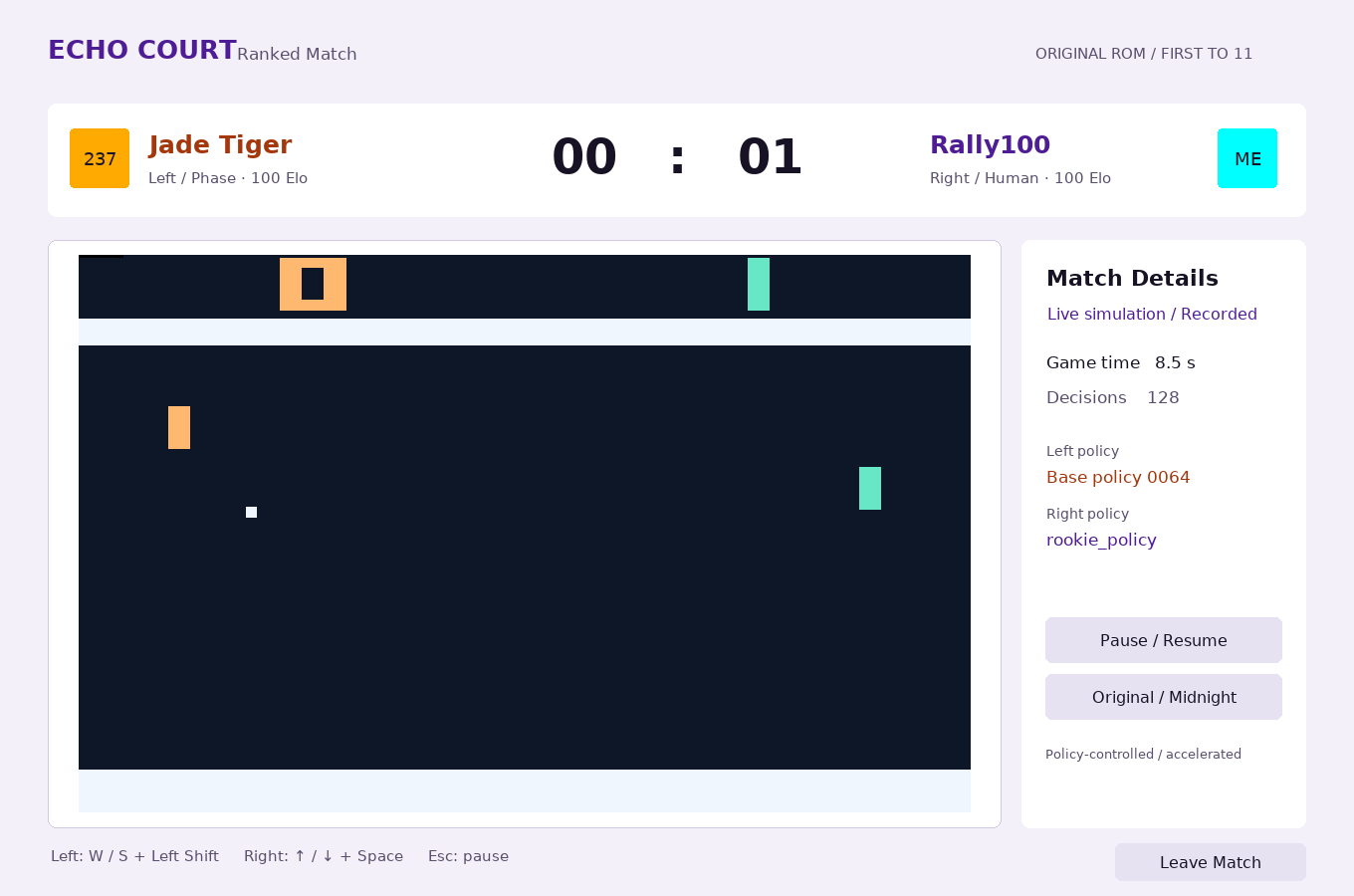}}
\caption{An actual match in the 50-match fixed-policy experience between Rally100, which uses the unoptimized initial program from the reference commit to control the right paddle, and a composite player; screenshot produced by off-screen rendering of the actual pygame window.}\label{fig:echo-court-match}
\end{figure}

The following two screenshots, Figures~\ref{fig:echo-court-entry} and~\ref{fig:echo-court-match}, were produced by off-screen rendering of the actual pygame window.

\section{Direct LLM Decision Baseline}\label{direct-llm-json-decision-baseline}
\subsection{Execution Protocol}\label{e.1-execution-protocol}

To demonstrate the capability of the model teacher, we used a more aggressive scheme, using an LLM, GLM-5.3, to output structured instructions and using it to directly control the single-step model. The LLM uses the same model-request configuration as the original learning runs. At every environment step, it reads the current state and outputs a JSON action. The executor advances the environment after validating the response; an invalid response triggers another request, and the environment-step count increases only after a legal action is executed. The model continues to make decisions directly throughout this process, with zero within-task training episodes. One episode was ultimately completed on test-evaluation seed 10000 for CartPole-v1.

\subsection{Score and Inference Cost}\label{e.2-score-and-inference-cost}

\begin{table}[t]
\centering
{\small
\caption{Return and execution investment for the single successful CartPole-v1 episode of the direct JSON-decision baseline. Request latencies come from usage records; extrema and the median follow the experiment report.}\label{tab:e1-json-episode}
\par\smallskip
\setlength{\tabcolsep}{4pt}
\begin{tabular}{@{} 
  >{\raggedright\arraybackslash}p{(\linewidth - 2\tabcolsep) * \real{0.55}}
  >{\raggedleft\arraybackslash}p{(\linewidth - 2\tabcolsep) * \real{0.45}}@{}}
\toprule
\begin{minipage}[b]{\linewidth}\raggedright
Metric
\end{minipage} & \begin{minipage}[b]{\linewidth}\raggedleft
Result
\end{minipage} \\
\midrule
CartPole return & 500.000 \\
Input tokens & 212633 \\
Output tokens & 3342490 \\
Cumulative model-request latency (h) & 9.377 \\
\bottomrule
\end{tabular}}
\end{table}

The episode remains balanced through the 500-step limit and reaches the maximum CartPole return. Output tokens are used primarily for model inference; under a protocol that waits for an action at every step, cumulative request latency becomes execution time. If 30 episodes each execute 500 steps with the same request overhead as this episode, linear extrapolation gives approximately 281.308 cumulative hours to execute this task. This is a budget estimate under the stated conditions.

This result reveals the powerful capability of general-purpose LLMs, far stronger than Jev-like models. However, due to time constraints, it is impossible to fully use such System 2 LLMs for inference. This is precisely the reason we turned the system into System 0.

\section{Details of Model and Ablation Results}\label{heatmap-analysis}

\subsection{Model Heatmap}\label{model-task-heatmap-appendix}

\begin{figure}
\centering
\pandocbounded{\includegraphics[width=\linewidth,keepaspectratio]{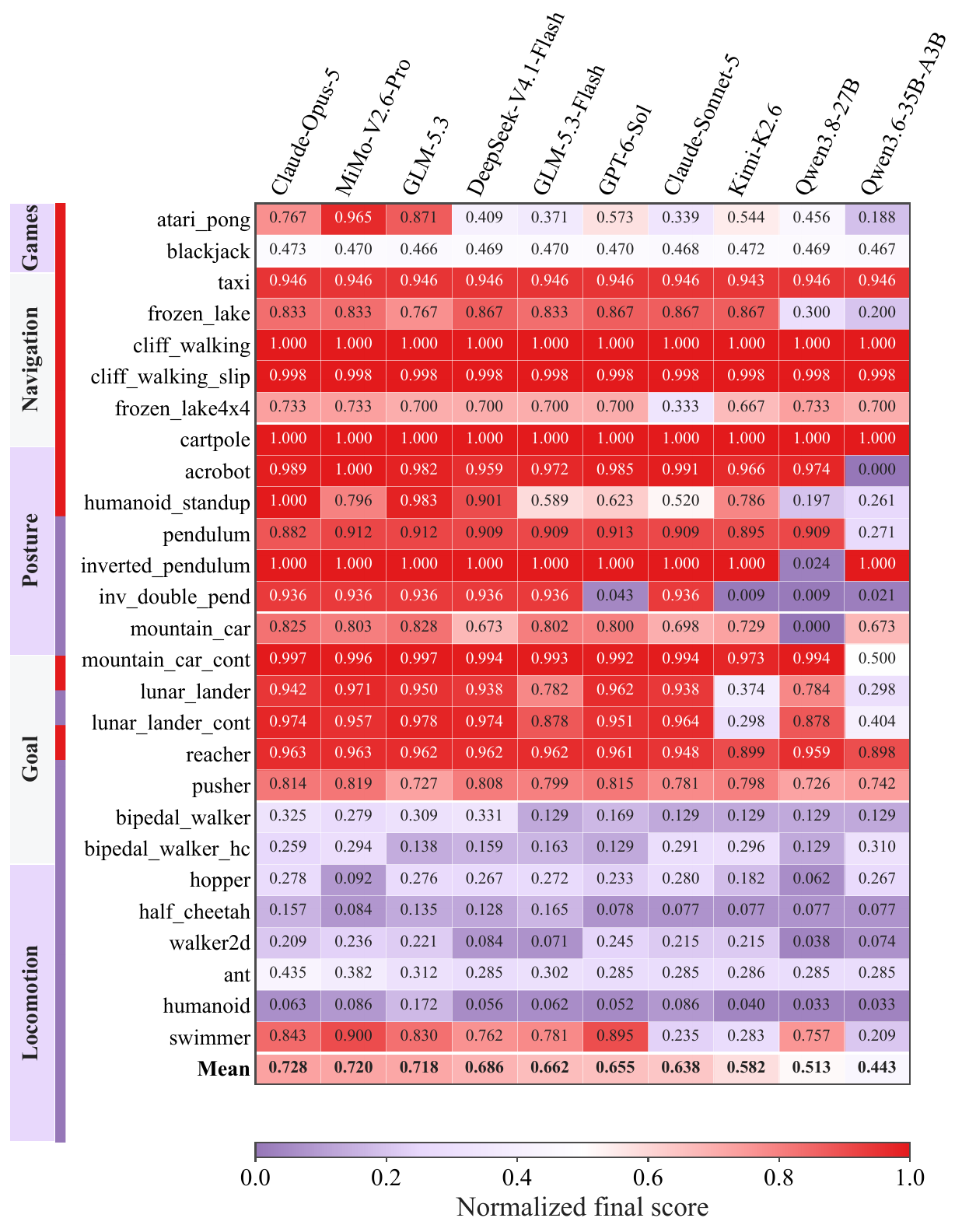}}
\caption{Test normalized scores for ten models across 27 tasks. Rows are grouped into five semantic task categories, the bar on the left distinguishes discrete and continuous actions, and model means appear at the bottom. The matrix supports comparison of model-task differences and action-type groups. Red indicates higher scores and blue lower scores; each cell is labeled to three decimal places.}
\label{fig:model-task-heatmap}
\end{figure}

Figure~\ref{fig:model-task-heatmap} presents per-task test normalized scores for the ten models across 27 tasks. Rows are grouped into five semantic task categories, and the bar on the left distinguishes discrete and continuous actions. The matrix supports comparison of model-task differences and action-type groups; model means appear at the bottom.

\subsection{Ablation Heatmap}\label{complete-results-of-ablation-experiments}

\begin{figure}[t]
\centering
\pandocbounded{\includegraphics[width=\linewidth,keepaspectratio]{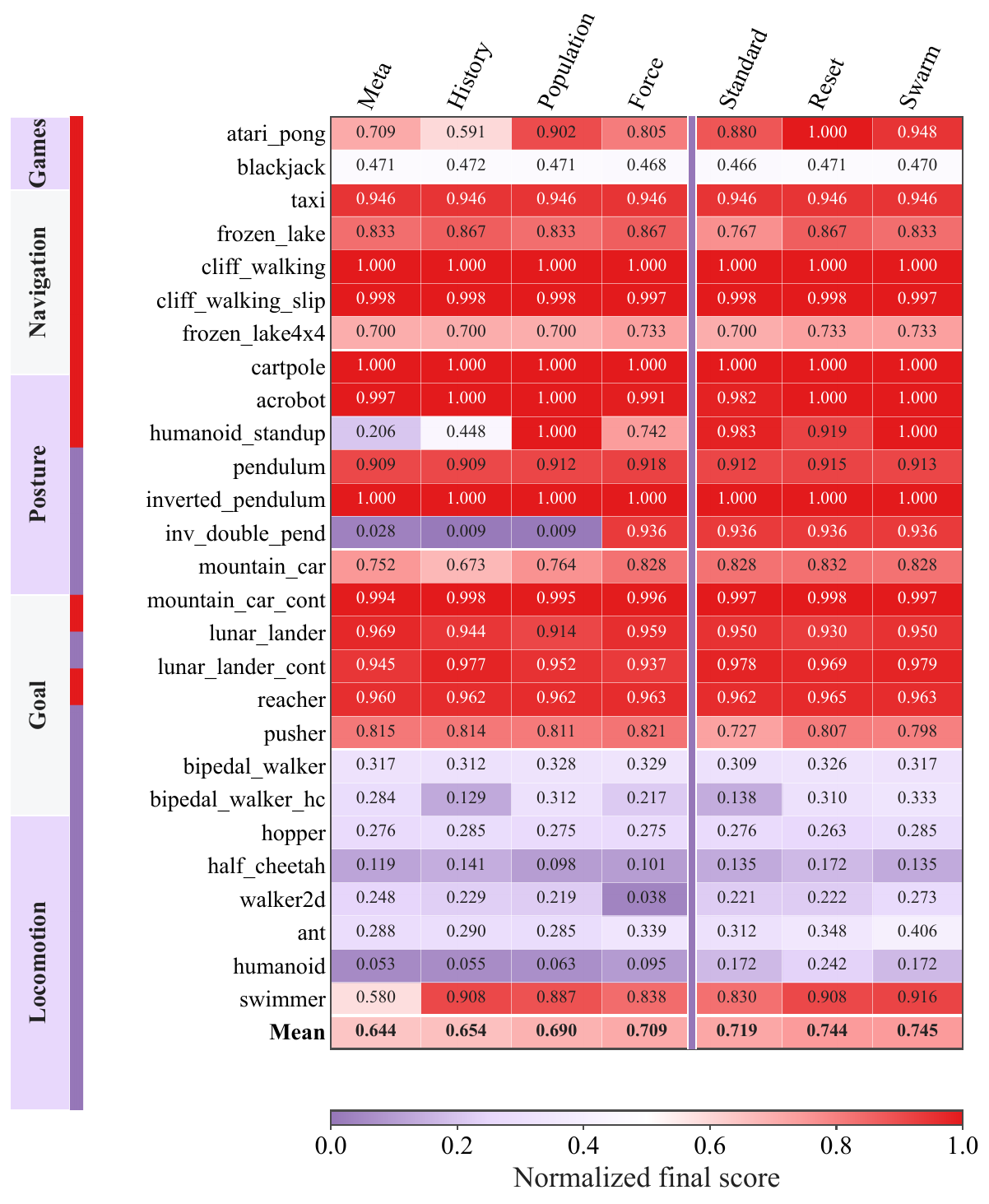}}
\caption{The ablation experiments are executed on the normalized scores on each task in the test set, and the bottom shows the average of each method across 27 tasks.}\label{fig:ablation-task-heatmap}
\end{figure}

Figure~\ref{fig:ablation-task-heatmap} reports complete results for all ablation experiments, including the individual results for the Meta, History, Population, and Force categories, as well as Standard, Reset, and Swarm for PDMetaHarness. Interestingly, although these methods do not surpass our Standard method in overall performance, they are actually somewhat complementary to one another. If we select the best-performing methods based on development-set performance and combine these methods, the average can still reach 0.742, which is slightly inferior to the 0.746 of the PDMetaHarness swarm method, but still a large improvement over the original baseline of 0.718. Testing and using more strategies may be a key direction for future research.

\section{Environment Seed Setting}\label{environment-seed-setting}

\paragraph{Basic Rules}

If a suitable evaluation score requires running $n$ times, then for all models, the seeds used for development are $0$ to $n$, and the seeds used for testing are $10000$ to $10000+n$.

\paragraph{Failed Seed Replacement Rules}

There are some special cases. That is, the environment may crash because it takes too long or falls into an infinite loop, with Atari-pong being a typical example. This is not a problem with the policy itself, but with the environment setup. We replace seeds where this occurs according to specific mapping rules. Specifically, the run result of the $k$-th failed seed is replaced by the result of $100000+k$. For example: in the History supplementary evaluation, the failed environment seeds 10016, 10020, and 10027 for Atari-pong were replaced by 100000, 100001, and 100002. Backup seeds are assigned independently according to each policy's own failure evidence and a fixed order.

\section{Token Costs}\label{token-cost-calculation-and-comparison-conditions}

We aggregate the input and output token count fields separately for each task. All 10 models have 27 trajectories with computable durations. The token counts reflect the usage recorded by the adapter. We did not calculate monetary cost because it depends on model pricing and cache hit rates. The results show that Claude-Opus-5 used the most candidate rounds, 645, while MiMo-V2.6-Pro had the highest consumption of both input and output tokens.

\begin{table}[t]
\centering
{\small
\setlength{\tabcolsep}{2pt}
\caption{Training token and round investment of ten models in selected sessions. Token units are in millions.}\label{tab:token-cost}
\par\smallskip
\begin{tabular}{@{}lccc@{}}
\toprule
Model & \shortstack{Candidate\\Rounds} & \shortstack{Input tokens\\(M)} & \shortstack{Output tokens\\(M)} \\
\midrule
Claude-Opus-5 & 645 & 86.421 & 1.897 \\
MiMo-V2.6-Pro & 404 & 449.347 & 7.091 \\
GLM-5.3 & 453 & 385.133 & 4.052 \\
DeepSeek-V4.1-Flash & 457 & 63.432 & 1.533 \\
GLM-5.3-Flash & 324 & 235.451 & 3.789 \\
GPT-6-Sol & 512 & 61.230 & 0.433 \\
Claude-Sonnet-5 & 425 & 211.916 & 2.584 \\
Kimi-K2.6 & 332 & 113.953 & 3.992 \\
Qwen3.8-27B & 78 & 59.154 & 2.123 \\
Qwen3.6-35B-A3B & 92 & 179.409 & 1.331 \\
\bottomrule
\end{tabular}}

\end{table}

\section{System 0/1/2/3}

The System 0/1/2/3 framework adopted in this paper is based on the quad-process cognitive framework~\cite{taniguchi2025quad}. It extends the conventional fast/slow distinction of dual-process theory to four temporal scales: super-fast, fast, slow, and super-slow. Each scale corresponds to a distinct class of cognitive processes, as summarized in Table~\ref{tab:system0123}.

\textbf{System 0} encompasses pre-cognitive embodied processes that operate on a super-fast timescale. These processes arise directly from physical interactions between an agent's body and its environment, with morphological computation and passive dynamics serving as their primary mechanisms.

\textbf{System 1} encompasses fast, intuitive, and automatic processes associated with world models and implicit representations. It primarily relies on tacit knowledge that is difficult to verbalize, enabling subconscious pattern recognition and immediate responses. Such processes operate mainly at the individual level.

\textbf{System 2} encompasses slow, deliberate, and effortful reasoning based on language models and explicit internal representations. It supports conscious, step-by-step reasoning and enables the organization of thought through inner language. Such processes operate mainly at the personal level.

\textbf{System 3} encompasses collective intelligence and symbol emergence on a super-slow timescale. Its central mechanism is collective prediction, through which multiple agents collaboratively construct and refine a shared world model via predictive processing. This process enables external symbolic systems, such as language, cultural norms, and social institutions, to emerge and evolve over time.

\begin{table}[t]
\centering
{\small
\caption{Key characteristics of the System 0/1/2/3 framework.}
\label{tab:system0123}
\par\smallskip
\setlength{\tabcolsep}{4pt}
\begin{tabular}{@{} 
  >{\raggedright\arraybackslash}p{(\linewidth - 8\tabcolsep) * \real{0.22}}
  >{\raggedright\arraybackslash}p{(\linewidth - 8\tabcolsep) * \real{0.195}}
  >{\raggedright\arraybackslash}p{(\linewidth - 8\tabcolsep) * \real{0.195}}
  >{\raggedright\arraybackslash}p{(\linewidth - 8\tabcolsep) * \real{0.195}}
  >{\raggedright\arraybackslash}p{(\linewidth - 8\tabcolsep) * \real{0.195}}@{}}
\toprule
Dimension & \textbf{System 0} & \textbf{System 1} & \textbf{System 2} & \textbf{System 3} \\
\midrule
Temporal scale & Super-fast & Fast & Slow & Super-slow \\
Underlying model & Physical dynamics & World model & Language model & Emergent symbolic system \\
Representation & Pre-representational & Implicit & Explicit and internal & External and shared \\
Cognitive process & Reflexive & Intuitive & Deliberate & Collective \\
Role of language & Non-linguistic & Sub-linguistic & Inner language & External language \\
\bottomrule
\end{tabular}}
\end{table}

Within this framework, we classify chain-of-thought reasoning in LLMs as System 2. Jev-like models and classical reinforcement learning methods are classified as System 1 because they rely on implicit representations and relatively simple model-based intuitive inference. Finally, executable programs obtained through program distillation are classified as System 0 because their execution requires no online model inference.

\section{Definition and Distinction of Program Distillation}

\subsection{Program Distillation versus Knowledge Distillation}

It should be noted that, although we use the term \emph{distillation}, program distillation (PD), as defined in this paper, is not identical to knowledge distillation in the conventional sense. PD refers to the transfer of capabilities from System 2 to System 0, whereas conventional knowledge distillation typically trains a student to match a teacher's logits or soft labels on observed inputs. The key distinction is that the student in PD is a program, and its learning does not rely on soft labels. Instead, the student inherits a programmatic representation in which the teacher's knowledge for solving the control problem has been encoded. Thus, PD differs from conventional knowledge distillation in both the representation of knowledge and the mechanism by which that knowledge is transferred. Nevertheless, the two share the same underlying objective: compressing the decision tendencies of a strong teacher into the behavior of a weaker student.

This perspective on program distillation, together with the introduction of teacher alignment, is essential for understanding our approach. Unlike conventional distillation, the direction of model updates in PD is not determined by gradients. Instead, updates are restricted to the hypothesis space and behavioral priors specified by the teacher. Furthermore, the student does not search for a solution from scratch. Its capabilities do not emerge solely through independent discovery based on task feedback. Rather, they are transferred directly from the teacher through a programmatic representation of the teacher's knowledge. During deployment, the student also inherits the teacher's action style in the form of an executable program, as demonstrated by our style distillation strategy.

\subsection{Formal Differences from Related Work}\label{app:distill-opt}

Recent studies often formulate code as a policy and optimize it directly. While related to our setting, this line of research diverges from our focus in a fundamental way. Rather than attributing performance gains solely to the optimization procedure, our study traces the observed performance gap to the underlying capability that generates the policy. Our core claim is that distillation and optimization serve as coexisting sources of capability, meaning that neither can be overlooked when accounting for simulation performance.

A formal comparison clarifies this distinction. Existing approaches can be formulated as
\begin{equation}
\mathcal O_\theta=(M_\theta,H_{\mathrm{special}}),\qquad
S'_\phi=\mathsf{Learn}(S_\phi,E,C;\mathcal O_\theta).
\end{equation}
where $M_\theta$ denotes the base model, $H_{\mathrm{special}}$ is a task-specific harness, $\mathcal O_\theta$ is the resulting optimization operator, $S_\phi$ represents the student policy, $E$ and $C$ denote the environment and context, and $\mathsf{Learn}$ is the optimization algorithm. Under this paradigm, policy capability stems primarily from optimization guided by a bespoke harness.

In contrast, our formulation introduces an explicit teacher:
\begin{equation}
(T_\theta,\mathcal O_\theta)=\mathcal K(M_\theta,H),\qquad
S'_\phi=\mathsf{Distill}(S_\phi,T_\theta,E,C;\mathcal O_\theta).
\end{equation}
where $H$ represents general knowledge regarding the target behavior, and the operator $\mathcal K$ constructs both the teacher $T_\theta$ and the optimization operator $\mathcal O_\theta$ from the base model and this knowledge. The improved student policy $S'_\phi$ is subsequently obtained through joint distillation and optimization. This formulation departs from prior work in two key respects. First, it assigns an explicit role to the teacher, requiring the base model to be structured into a source of transferable behavioral knowledge. Second, it provides a generalizable procedure that eliminates the reliance on a manually engineered, task-specific harness such as $H_{\mathrm{special}}$, which is typically required by code-as-policy methods such as Evolution of Heuristics \cite{liu2024eoh}. Consequently, distillation and optimization act as complementary sources of capability, and our analysis emphasizes the former just as much as the latter.

\end{document}